\documentclass[10pt]{article} 
\usepackage[accepted]{tmlr}

\usepackage{amsmath,amsfonts,bm}

\def\eqref#1{equation~\ref{#1}}

\def\1{\bm{1}}

\DeclareMathAlphabet{\mathsfit}{\encodingdefault}{\sfdefault}{m}{sl}
\SetMathAlphabet{\mathsfit}{bold}{\encodingdefault}{\sfdefault}{bx}{n}

\usepackage[colorlinks=true,citecolor=RoyalBlue,linkcolor=Bittersweet]{hyperref}
\usepackage{url}
\usepackage[dvipsnames,table]{xcolor}
\usepackage{pifont}
\usepackage{booktabs}
\usepackage{siunitx}
\usepackage{algorithm}
\usepackage{algpseudocode}
\algrenewcommand\algorithmicrequire{\textbf{Input:}}
\algrenewcommand\algorithmicensure{\textbf{Output:}}
\usepackage{algpseudocode}
\usepackage{threeparttable}
\usepackage{multirow}
\usepackage{tabularx}
\usepackage{cleveref}
\usepackage{enumitem}
\usepackage{mathtools}
\usepackage{soul}
\usepackage{multirow}
\usepackage{colortbl}
\usepackage{xspace}
\usepackage{subcaption}
\usepackage[most]{tcolorbox} 

\usepackage{listings}

\definecolor{codeblue}{rgb}{0.0,0.0,0.6}
\definecolor{codegreen}{rgb}{0.0,0.5,0.0}
\definecolor{codeorange}{rgb}{0.8,0.5,0.0}

\lstdefinestyle{pythonstyle}{
    language=Python,
    backgroundcolor=\color{white},   
    commentstyle=\color{codegreen},  
    keywordstyle=\color{codeblue},   
    stringstyle=\color{codegreen},   
    basicstyle=\ttfamily\small,      
    morekeywords={str, int, float, List}, 
    keywordstyle=[2]\color{codeorange},   
    breakatwhitespace=true,          
    breaklines=true,                 
    captionpos=b,                    
    frame=single,                    
    rulecolor=\color{black},         
    keepspaces=true,                 
    showstringspaces=false,          
    tabsize=4,                       
    xleftmargin=1em                 
}

\crefname{figure}{Figure}{Figures}
\crefname{section}{\S}{\S\S}
\crefname{equation}{Equation}{Equationns}
\crefname{appendix}{Appendix}{Appendices}
\crefname{table}{Table}{Tables}

\newcommand{\think}[1]{\textcolor{blue}{\texttt{<think>}}#1\textcolor{blue}{\texttt{</think>}}}
\newcommand{\search}[1]{\textcolor{cyan}{\texttt{<search>}}#1\textcolor{cyan}{\texttt{</search>}}}
\newcommand{\info}[1]{\textcolor{brown}{\texttt{<information>}}#1\textcolor{brown}{\texttt{</information>}}}
\newcommand{\answer}[1]{\textcolor{purple}{\texttt{<answer>}}#1\textcolor{purple}{\texttt{</answer>}}}
\newcommand{\ours}{\textsc{VERITAS}\xspace}
\newcommand{\veritasrone}{\textsc{VERITAS-R1}\xspace}
\newcommand{\research}{\textsc{ReSearch}\xspace}
\newcommand{\searchrone}{\textsc{Search-R1}\xspace}
\newcommand{\desa}{\textsc{DeSA}\xspace}

\title{Beyond Correctness: Rewarding Faithful Reasoning\\in Retrieval-Augmented Generation}

\author{\noindent\name Zhichao Xu$^{1}$\thanks{Both authors contributed equally to this work.} \email xzhichao@amazon.com \vspace{3pt}\\
\name Zongyu Wu$^{2}$\footnotemark[1] \thanks{Work done during an internship at Amazon Web Services.} \email zongyuwu@psu.edu \vspace{3pt}\\
\name Yun Zhou$^{1}$ \vspace{3pt}\\
\name Aosong Feng$^{1}$ \vspace{3pt}\\
\name Kang Zhou$^{1}$ \vspace{3pt}\\
\name Sangmin Woo$^{1}$ \vspace{3pt}\\
\name Kiran Ramnath$^{1}$ \vspace{3pt}\\
\name Yijun Tian$^{1}$ \vspace{3pt}\\
\name Xuan Qi$^{1}$ \vspace{3pt}\\
\name Weikang Qiu$^{3}$\footnotemark[2] \vspace{3pt}\\
\name Lin Lee Cheong$^{1}$ \vspace{3pt}\\
\name Haibo Ding$^{1}$ \email hbding@amazon.com \vspace{3pt}\\
\addr $^{1}$AWS AI Fundamental Research \hspace{1em} $^{2}$The Pennsylvania State University\hspace{1em} $^{3}$Yale University
}

\def\month{05}  
\def\year{2026} 
\def\openreview{\url{https://openreview.net/forum?id=mZ0gGlXelF}} 

\begin{document}

\maketitle

\begin{abstract}
Inspired by the success of reinforcement learning (RL) in Large Language Model (LLM) training for domains like math and code, recent work has begun training LLMs to dynamically plan, query, and reason with search engines as tools\,---\,a paradigm increasingly referred to as \textit{agentic search}. Although these methods achieve performance improvement across popular short-form QA benchmarks, many prioritize final answer correctness while overlooking the quality of intermediate reasoning steps, which may lead to \textit{chain-of-thought unfaithfulness}. In this paper, we first introduce a comprehensive evaluation framework for agentic search, covering three distinct faithfulness metrics: \textit{Think-Search faithfulness}, \textit{Information-Think faithfulness}, and \textit{Think-Answer faithfulness}. Our evaluations reveal that canonical agentic search systems trained through Reinforcement Learning from Verifiable Reward (RLVR) using episode-level outcome-based reward\,---\,including \searchrone and \research---\,have significant room for improvement on these faithfulness dimensions. To foster faithful reasoning in agentic search, we introduce \ours (Verifying Entailed Reasoning through Intermediate Traceability in Agentic Search), a novel framework that integrates fine-grained turn-level faithfulness rewards into the reinforcement learning process. Our experiments show that models trained with \ours not only significantly improve reasoning faithfulness, but also achieve better task performance compared to baselines trained against episode-level outcome-based reward.
\end{abstract}

\section{Introduction}
\label{sec:introduction}
Large Language Models (LLMs) have achieved remarkable success across diverse domains~\citep{brown2020language,google2024gemini,grattafiori2024llama3}, yet they continue to face persistent challenges such as hallucinations~\citep{li2023halueval,maynez-etal-2020-faithfulness,Huang2025SurveyOnHallucination} and outdated knowledge~\citep{mousavi2024dyknow}. While Retrieval-Augmented Generation~\citep[RAG,][]{Patrick2020rag,zhao2024retrieversurvey} mitigates these issues by grounding outputs in external knowledge, conventional retrieve-then-generate pipelines often fail on complex reasoning queries that require multi-turn interaction and evidence synthesis~\citep{gao2023ragsurvey,trivedi2023ircot,yao2023react}. This has motivated a shift toward \textit{agentic search}~\citep{singh2025agentic,liang2025reasoningrag}\,---\,a paradigm in which LLMs are trained via reinforcement learning to dynamically plan, issue queries to a retriever, and reason over returned evidence as part of an interactive trajectory~\citep{jin2025searchr1,li2025searcho1,chen2025research,song2025r1searcher}.

Despite impressive benchmark results, existing agentic search methods still face a critical limitation: they optimize for final-answer correctness~\citep{jin2025searchr1,chen2025research,li2025searcho1} while overlooking the faithfulness of intermediate reasoning steps~\citep{baker2025monitoring,bao-etal-2025-howlikely}. As recent work reveals~\citep{Lanham2023MeasuringCotFaith,bentham2024chainofthoughtunfaithfulness}, outcome-based training often produces reasoning traces that misalign with the final answer, a phenomenon termed as \textit{chain-of-thought unfaithfulness}.
Maintaining Chain-of-Thought (CoT) faithfulness~\citep{Lanham2023MeasuringCotFaith,baker2025monitoring} and contextual faithfulness~\citep{malaviya2025contextualized} are critical for reasoning models, particularly for RL-based search agents, as the intermediate CoT steps provide users with a clear reasoning pathway to understand how a complex query is decomposed into atomic queries used for retrieving evidence and leads to the final answer that addresses the advanced information need~\citep{marchionini2006exploratory}. However, it remains unclear how reasoning faithfulness in the context of agentic search should be formulated and evaluated. 

\begin{figure*}[t]
    \centering
    \includegraphics[width=\textwidth]{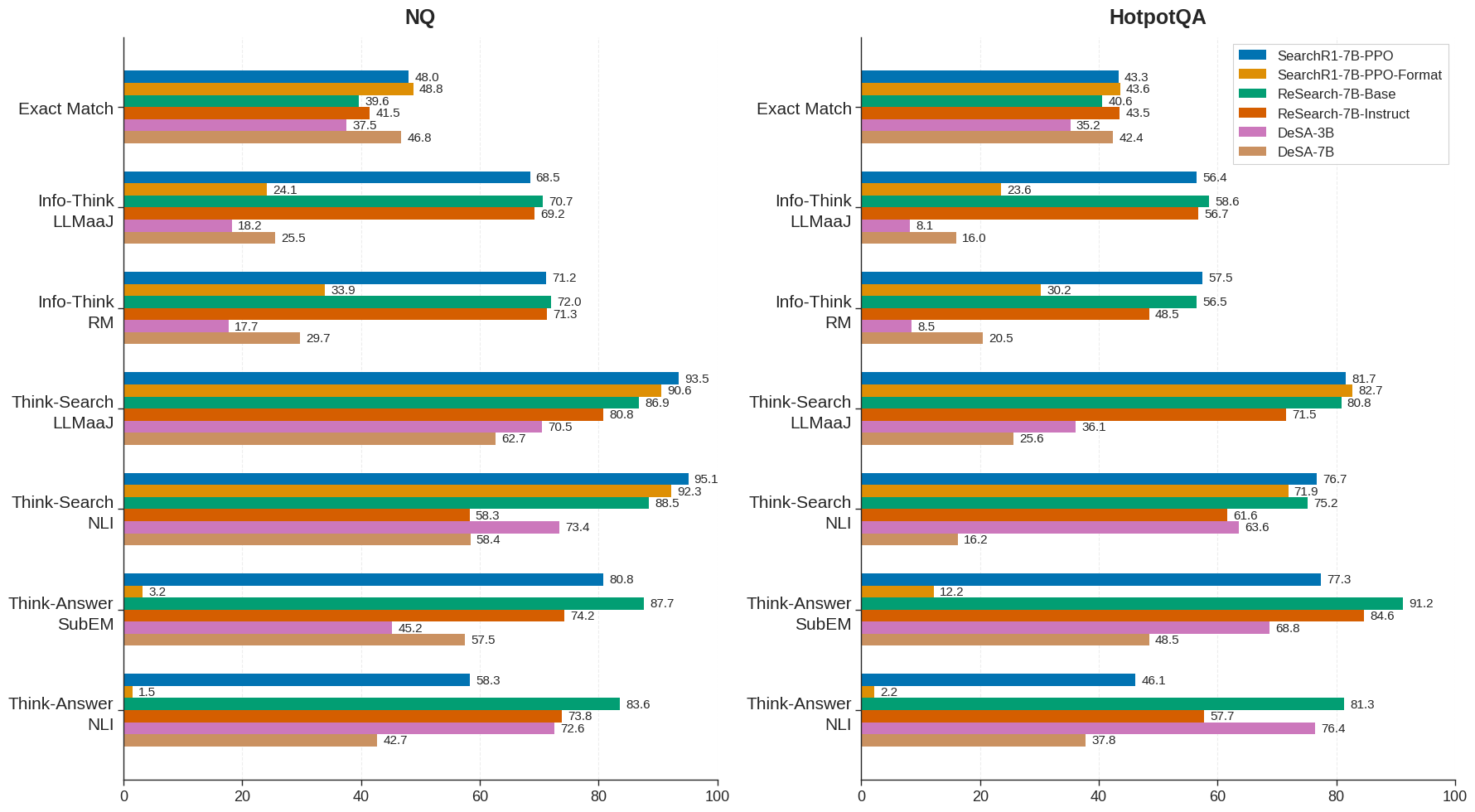}
    \vspace{-10pt}
    \caption{Evaluation of \searchrone, \research and \desa on task performance (Exact Match) and faithfulness dimensions. Left: results on NQ dataset. Right: results on HotpotQA dataset. We can see models trained with pure outcome-based reward, their good task performance does not automatically guarantee fully faithful reasoning chains --- notably the Information-Think faithfulness dimension --- which may be suboptimal for the practical deployment of such systems.}
    \label{fig:search_r1_faithfulness_eval}
\end{figure*}

To systematically address this ambiguity, we first formalize and quantify faithfulness in agentic search. We propose a novel evaluation framework centered on three key dimensions of reasoning traceability: \textbf{Think-Search} faithfulness (\textit{are search queries justified by the agent's thoughts?}), \textbf{Information-Think} faithfulness (\textit{do the agent's thoughts consider the retrieved information?}), and \textbf{Think-Answer} faithfulness (\textit{is the final answer grounded in the agent's thoughts?}). Applying this framework to popular agentic search models such as \searchrone~\citep{jin2025searchr1} and \research~\citep{chen2025research}, reveals a critical gap: as shown in~\cref{fig:search_r1_faithfulness_eval}, the gain in the final-answer accuracy does not automatically guarantee the faithfulness of intermediate reasoning steps.

Motivated by recent work showing that augmenting episode-level outcome rewards with feedback on intermediate reasoning can improve RL training~\citep{cobbe2021trainingverifiers,lightman2023letsverify,zhang2025lessonsofdeveloping}, we hypothesize that optimizing only outcome-based rewards while overlooking the quality of intermediate reasoning steps may hinder model performance. However, applying faithfulness as a training signal in agentic search is non-trivial: a single trajectory contains heterogeneous output types (thoughts, queries, retrieved evidence, answers), each of which requires a different notion of correctness. We therefore introduce \textbf{\ours} (\textbf{V}erifying \textbf{E}ntailed \textbf{R}easoning through \textbf{I}ntermediate \textbf{T}raceability in \textbf{A}gentic \textbf{S}earch), a framework designed to train verifiably faithful search agents. The name, derived from the Latin word for ``truth'' (\textit{vēritās}), reflects our objective to ensure the agent's reasoning is true to the evidence it gathers. \ours operationalizes this ideology by incorporating our proposed faithfulness metrics directly into the RL training loop as fine-grained, structured faithfulness rewards.\footnote{We use the terms \textit{structured faithfulness rewards} or \textit{checklist-style process rewards} throughout the paper, in contrast to the per-step process reward models of \citet{lightman2023letsverify} and \citet{cobbe2021trainingverifiers}, which score every reasoning step. \ours instead defines a small fixed set of faithfulness criteria tied to specific transitions in the agentic search trajectory (think$\to$search, information$\to$think, think$\to$answer), which is more closely related to recent rubric- and checklist-based evaluation~\citep{ribeiro-etal-2020-beyond,arora2025healthbenchevaluatinglargelanguage,deshpande-etal-2025-multichallenge} and rubric-as-reward RL~\citep{gunjal2026rubricsasrewards,viswanathan2025checklists,shao2025drtulureinforcementlearning}. All three lines of work share the goal of contrasting with episode-level outcome rewards, but differ in granularity and structure. We discuss this connection in detail in~\cref{sec:conclusion}.} Our trained model, \veritasrone, not only improves information-think faithfulness by around 14\% and think-answer faithfulness by around 7.7\% over its comparable \searchrone baseline, but also increases task accuracy.

In summary, our contributions are:
\begin{itemize}[leftmargin=*,itemsep=2pt,topsep=2pt]
\item We propose a formal definition and a corresponding evaluation framework for faithfulness in agentic search, centered on three novel metrics: Information-Think, Think-Answer, and Think-Search Faithfulness.
\item Through this framework, we conduct a comprehensive analysis of leading RL-based search agents, revealing a gap between their task performance and the reasoning faithfulness.
\item We introduce \ours, a novel training framework that integrates fine-grained faithfulness metrics as structured, checklist-style rewards into the RL loop. Our experiments show that the proposed method effectively improves the model’s faithfulness while maintaining comparable task accuracy across seven downstream QA benchmarks.
\end{itemize}

\section{Related Work}
\label{sec:related_works}

\noindent
\textbf{Faithfulness in reasoning and contextualization.}
Although explicit reasoning can enhance LLM performance, the presence of intermediate steps does not necessarily reveal the true causal mechanisms behind model outputs~\citep{turpin2023language,Lanham2023MeasuringCotFaith,balasubramanian2025cotbias}. A key question is whether these steps genuinely support the final answer or merely serve as plausible post-hoc explanations. This concern has motivated a growing body of research on \textit{Chain-of-Thought (CoT) faithfulness}~\citep{atanasova2023counterfactualfaithfulness,paul-etal-2024-making,balasubramanian2025cotbias,chua2025deepseekfaith,parcalabescu2024measuring}.  
Counterfactual approaches~\citep{atanasova2023counterfactualfaithfulness,chen2023counterfactualcot} analyze how perturbing reasoning steps affects model predictions, while adversarial methods~\citep{Lanham2023MeasuringCotFaith,matton2025faithmeasure} intervene on the CoT or model states to probe causal dependencies~\citep{balasubramanian2025cotbias}.  
Another popular direction focuses on bias articulation and explanation reliability~\citep{turpin2023language,balasubramanian2025cotbias,arcuschin2025cotfaitheval}.  

Beyond reasoning traces, LLMs often exhibit \textit{contextual unfaithfulness}\,---\,producing outputs that contradict or are unsupported by given contexts~\citep{Huang2025SurveyOnHallucination,bi2024context,ye2023cognitivehallusurvey}. This phenomenon limits their reliability in applications requiring factual grounding. Prior studies~\citep{dziri2022origincontextfaith,dziri2022faithdial,ming2025faitheval,malaviya2025contextualized} have evaluated whether model responses are properly entailed by reference passages, emphasizing accurate synthesis without introducing unsupported claims.  
Together, these works reveal the importance of verifying both reasoning and contextual grounding, yet they primarily focus on static text generation rather than dynamic, search-driven reasoning.

\noindent
\textbf{Retrieval-Augmented Generation.}
Retrieval-Augmented Generation~\citep[RAG,][]{Patrick2020rag,gao2023ragsurvey,jiang2023activerag,trivedi2023ircot,fan2024survey,chan2024rqrag,li2025searcho1,jiang2025ragstar,chang2025mainrag,xu2025reconreasoningcondensationefficient} enables LLMs to access external knowledge for more factual and up-to-date responses.  
RAG techniques span diverse directions~\citep{gao2023ragsurvey}, including query rewriting~\citep{ma2023queryRewriting,zheng2024takebackrag,xu2025rethinkingonpolicyoptimizationquery}, iterative retrieval~\citep{li2025searcho1}, and supervised fine-tuning for tool use~\citep{schick2023toolformer,asai2023selfrag,patil2024gorilla}.  
Despite their effectiveness on QA and knowledge-intensive tasks~\citep{Patrick2020rag,xu2025chatqav2}, many RAG systems depend heavily on prompt engineering or are difficult to scale~\citep{sun2025zerosearch}.  
More critically, they are typically evaluated on final-answer accuracy, overlooking whether the intermediate reasoning steps remain faithful to the retrieved evidence.

\noindent
\textbf{Agentic Search with Reinforcement Learning.}
Reinforcement Learning~\citep[RL,][]{watkins1992qlearning,kaelbling1996reinforcement,sutton1998reinforcement} has been widely adopted to align model behavior with long-term objectives.  
Early work explored RL from human feedback~\citep{stiennon2020learningtosummarize,ouyang2022training,pmlr-v238-gheshlaghi-azar24a,kaufmann2025a}, often using Proximal Policy Optimization~\citep[PPO,][]{schulman2017proximal}.  
Recent methods such as REINFORCE Leave-One-Out~\citep[RLOO,][]{kool2019buyfoursamples,ahmadian-etal-2024-back}, Group Relative Policy Optimization~\citep[GRPO,][]{shao2024deepseekmathgrpo,yu2026dapo} improve stability and sample efficiency by removing the need for a separate critic, while off-policy preference optimization methods~\citep{rafailov2023direct,meng2024simpo} are considered more scalable but lag behind in terms of performance~\citep{lanchantin2025bridging,lin2025comprehensivesurveyreinforcementlearningbased}.

Chain-of-Thought prompting~\citep{wei2022chain,kojima2022largelanguagemodelsarezeroshotreasoners} decomposes complex tasks into interpretable steps~\citep{Geva2021qareasonbenchmark,chu2024cotsurvey} to achieve better performance. ReAct~\citep{yao2023react} asks the LLM-based agent to reason before taking actions. Based on these technical approaches, recent works have explored training reasoning-focused LLMs~\citep{jaech2024openai}. For instance, DeepSeek-R1~\citep{guo2025deepseek} employs GRPO to generate long reasoning chains that solve challenging queries.
Inspired by this success, several works have applied RL to CoT-based reasoning, training LLMs to interact with search tools~\citep{chen2025research,jin2025searchr1,song2025r1searcher}.  
While outcome-based approaches~\citep{jin2025searchr1,li2025searcho1} optimize for final-answer correctness, others design intermediate rewards to encourage more deliberate search or reasoning~\citep{jin2025searchr1extension,wang2025stepsearch,shi2025autorefine,zhao2025rsearch}.  
For example, AutoRefine~\citep{shi2025autorefine} introduces a retrieval-specific reward that encourages models to utilize evidence effectively\,---\,a signal conceptually related to faithfulness.  

However, no prior work has comprehensively evaluated or formalized the notion of faithfulness in RL-based agentic search.  
To fill this gap, we propose an evaluation framework consisting of three distinct faithfulness dimensions for agentic search use case; and further proposes a training-based approach that integrates these faithfulness dimensions as fine-grained, structured rewards at key trajectory transitions.

\section{A Framework for Evaluating Faithfulness in Agentic Search}
\label{sec:evaluation}
As noted in~\cref{sec:introduction}, a critical limitation of current RL-based search agents is that the policy trained with pure outcome-based reward may elicit unfaithful reasoning. To systematically diagnose this issue, we need to establish a formal evaluation. We first brief the essential background, then introduce our proposed evaluation framework.

\subsection{Background: Agentic Search Trajectory}
\label{subsec:background}
In agentic search, LLMs dynamically use search engines as tools. We select \searchrone~\citep{jin2025searchr1} as a canonical example (algorithm details in~\cref{asec:ppo}). The interaction process produces a structured trajectory containing text blocks wrapped with four types of tags~\citep{jin2025searchr1}:
\begin{itemize}[leftmargin=*,nosep]
    \item \think{}: The agent's internal reasoning, where it analyzes information and plans its next step.
    \item \search{}: The search query generated by the agent to send to the retriever.
    \item \info{}: The retrieved information from the search tool.
    \item \answer{}: The agent's final answer to the initial query.
\end{itemize}
Since the RL-based search agent is purely trained to optimize the outcome-based reward (e.g., the correctness of the final answer), it overlooks the integrity of this intermediate trajectory. To address this gap, we develop a framework to systematically measure the faithfulness of the intermediate reasoning steps.

\subsection{Faithfulness Definitions}
\label{subsec:faithfulness_definition}
To systematically analyze the quality of the reasoning process in agentic search, we introduce a structured framework based on three distinct, turn-by-turn notions of faithfulness. These definitions provide a theoretical grounding for our subsequent evaluation. Let a trajectory be a sequence of alternating thoughts, search queries, and retrieved information, culminating in a final answer. We define faithfulness at three critical junctures within this process. Formally, we define the following three types of faithfulness in the context of agentic search:

\noindent
\textbf{Think-Search faithfulness.} 
This dimension assesses the logical connection between the agent's internal reasoning process and its subsequent search query. A search query is considered faithful if it is a direct and necessary consequence of an information gap explicitly identified or implicitly raised in the immediately preceding thought process. The core principle is that the agent should ``search what it thinks it needs'', ensuring that its information-gathering actions are deliberate and well-motivated by its reasoning state.

\noindent
\textbf{Information-Think faithfulness.} This dimension evaluates whether the agent's reasoning properly uses the evidence it has just acquired. The thinking block following an information block is considered faithful if the reasoning it contains is a valid synthesis, summary, or logical deduction based on the newly retrieved information. This form of faithfulness is crucial for preventing the model from ignoring evidence, thereby ensuring the integrity of each step in the reasoning chain.

\noindent
\textbf{Think-Answer faithfulness.} This dimension measures whether the final answer is supported by the cumulative knowledge gathered throughout the entire trajectory. The final answer is deemed faithful if the final claim it makes is entailed by the content within the \textit{latest preceding thinking block}. This formulation ensures that the agent does not introduce new, unverified information at the final step and that its conclusion is attributed to the complete, self-contained search and reasoning trajectory.

\subsection{Evaluation Metrics}
\label{subsec:evaluating_faithfulness}
To operationalize the faithfulness dimensions defined in \cref{subsec:faithfulness_definition}, we design evaluation metrics that align with the semantic role of each transition in the agentic search trajectory.
We combine lightweight automatic checks with LLM-based judgment, reflecting the different levels of semantic subtlety required across faithfulness types. Metrics are organized following the temporal order of the trajectory.

\begin{itemize}[leftmargin=*]
    \item \textbf{Think--Search Faithfulness.} Think--Search faithfulness assesses whether a search action is logically motivated by the immediately preceding reasoning. We first formulate this dimension as a textual entailment problem, treating the \think{} block as the \textit{premise} and the subsequent \search{} query as the \textit{hypothesis}. Using a pretrained NLI model, \texttt{google/t5\_xxl\_true\_nli\_mixture}~\citep{honovich-etal-2022-true-evaluating}, we label each (\think{}, \search{}) pair as Entailment or No entailment. Because strict entailment may undercount faithful searches that are implicitly motivated or differently abstracted, we additionally employ an LLM-as-a-Judge (LLMaaJ) evaluation. The judge LLM assesses whether the search query is a reasonable and well-motivated response to the agent’s reasoning, flagging irrelevant or disconnected queries. Unless otherwise specified, we use \texttt{Claude Sonnet-4.5} as the judge, with the prompt shown in Appendix~\cref{tab:think-search-template}. We report both NLI-based and LLMaaJ-based scores and analyze their agreement.
    \item \textbf{Information--Think Faithfulness.}
Information--Think faithfulness evaluates whether retrieved evidence is meaningfully incorporated into subsequent reasoning.
As this requires assessing synthesis and selective use of information, we adopt an LLM-as-a-Judge approach.
The judge is presented with (\info{}, \think{}) pairs and determines whether the reasoning is grounded in, consistent with, and responsive to the retrieved content, flagging ignored, contradicted, or unsupported claims.
The output is a binary judgment. The prompt template for judge model is provided in Appendix~\cref{tab:info-think-template}.
\item \textbf{Think--Answer Faithfulness.}
Think--Answer faithfulness measures whether the final answer is supported by the immediately preceding reasoning.
We operationalize this dimension using two automatic metrics.
First, we apply an NLI-based entailment check, treating the final \think{} block as the \textit{premise} and the \answer{} block as the \textit{hypothesis}.
Second, for short-form QA settings, we use a substring exact-match (Sub-EM) metric that verifies whether key entities or factual claims in the answer explicitly appear in the preceding reasoning.
We examine the agreement between these metrics in \cref{subsec:evaluating_searchr1_faithfulness} and discuss their respective trade-offs.
\end{itemize}

\begin{figure*}[t]
    \centering
    \includegraphics[width=\textwidth]{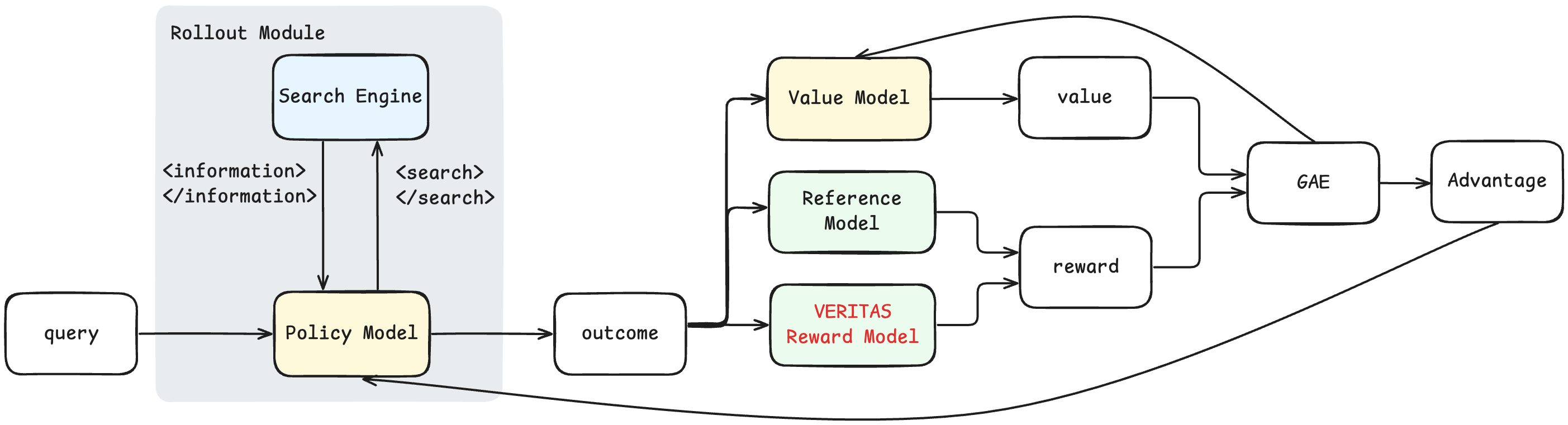}\vspace{-8pt}
    \caption{The pipeline of \ours-R1 in the RL framework with PPO RL algorithm. Applying process supervision (\textcolor{red}{\ours Reward Model}) improves the faithfulness of policy model's intermediate reasoning steps. Outcome in this figure refers to the rollout trajectory used for subsequent reward calculation and advantage estimation.}
    \label{fig:ours-pipeline}
\end{figure*}

\textbf{Generalizability.}
Currently, the proposed metrics are instantiated for \searchrone, but the underlying framework is broadly applicable to any reasoning-centric system that produces \textit{structured trajectories}, such as~\research~\citep{chen2025research} and \desa~\citep{wang2025outcomerewarddecouplingsearch}.
Because it operates over the abstract \textit{think–search–information–answer} schema, the evaluation methodology remains agnostic to model architecture and domain, which enables consistent assessment of turn-level faithfulness across diverse multi-turn retrieval and reasoning agents.  
This generalizability also allows the framework to serve as a foundation for developing structured, checklist-style training objectives, as demonstrated in the next section through the design of \ours.

\subsection{Faithfulness Evaluation Results}
\label{subsec:evaluating_searchr1_faithfulness}

To have a basic understanding of the faithfulness of existing agentic search models, we evaluate three RL-based RAG models: \searchrone~\citep{jin2025searchr1}, \research~\citep{chen2025research} and \desa~\citep{wang2025outcomerewarddecouplingsearch} trained against outcome-based rewards. We use their official checkpoints to ensure reproducibility. For Think-Search faithfulness and Think-Answer faithfulness, we also calculate the agreement rate and Fleiss's $\kappa$ as meta-evaluation of the two evaluation metrics.
From~\cref{fig:search_r1_faithfulness_eval}, we make these key observations. 
\begin{itemize}[leftmargin=*,nosep]
    \item \textbf{The Information-Think Faithfulness of evaluated models are far from perfect}. For example, \searchrone-7B-Base PPO has an Information-Think faithfulness score of 0.564 on the complex multi-hop HotpotQA dataset, while the best checkpoint we evaluated\,---\,\research-7B-Base achieves 0.586, suggesting that \textit{the policy models trained against pure outcome-based reward are not fully faithful to the retrieved information during the thinking process}.
    \item \textbf{Task performance does not directly translate to high faithfulness scores}. For example, \desa-7B achieves competitive performance on both NQ and HotpotQA datasets, but it consistently underperforms \searchrone and \research in almost all three faithfulness dimensions. 
    \item \textbf{All evaluated checkpoints have high Think-Search faithfulness}. For instance, \searchrone has a Think-Search score of 0.935 on NQ dataset when using LLMaaJ. The results suggest that their search query aligns well with their reasoning process. 
\end{itemize}
We also note that \searchrone-7B-Base-PPO-Format~\citep{jin2025searchr1extension} shows poor Information-Think faithfulness. A closer manual examination of its traces suggested that for the \think~ block after retrieved information, the policy often directly transitions to the next search query without reasoning over the documents, e.g., ``\info{...}\textcolor{blue}{\texttt{<think>}}I need to find out the name of the first person who got the Nobel Prize in Physics.\textcolor{blue}{\texttt{</think>}}''. The policy also tends to continue to search until reaching the maximum number of allowed search turns, then being forced to generate an answer, leading to poor Think-Answer faithfulness, e.g., ``\textcolor{blue}{\texttt{<think>}}I need to find out who got the first Nobel Prize in Physics.\textcolor{blue}{\texttt{</think>}}\textcolor{purple}{\texttt{<answer>}} Wilhelm Röntgen \textcolor{purple}{\texttt{</answer>}}''. These observations collectively reveal that \emph{a modified reward shaping by adding a format reward may improve model performance, but in return potentially breaks the model's reasoning consistency}.

\subsection{Meta-evaluation}
\label{subsec:meta-eval}

As we evaluate RL-based RAG models, we are also conducting meta evaluation of our evaluation metrics. We report detailed meta evaluation results in~\cref{tab:meta-eval}. While using a powerful LLM-as-a-Judge provides high-quality faithfulness labels, its cost and latency make it impractical for scalable evaluation. For Information-Think faithfulness, we propose to use a smaller reward model by distilling from the LLM judge, which we leave details to~\cref{sec:improve_searchr1}.  For Think-Search faithfulness, we observed high agreement rate between the LLM judge and the NLI model, thus we propose to use the NLI model as the automatic evaluation. For Think-Answer faithfulness, we notice a moderate inconsistency between Sub-EM and NLI metric. A closer manual evaluation suggests that as the RAG datasets primarily focus on short-form answers, the NLI model trained on symmetric (\texttt{Premise}, \texttt{Hypothesis}) pairs often predict ``No entailment'' for coherent think-answer pairs, leading to high False Positive rate. Therefore, we adopt Sub-EM as the automatic metric for Think-Answer faithfulness evaluation.

Our findings in this section highlight that, despite their strong answer accuracy, the three RL-based RAG models we examined still exhibit notable weaknesses in reasoning faithfulness\,---\,particularly in how retrieved evidence is integrated into the thought process. This observation underscores a fundamental limitation of existing agentic search models trained solely with outcome-based rewards: They fail to incentivize faithful reasoning traces to arrive at final answers.

\begin{table}[t!]
\centering
\caption{Meta-evaluation results on NQ and HotpotQA datasets. For \searchrone, we use the checkpoints using the Base model. For Think-Search and Think-Answer faithfulness, the Fleiss $\kappa$ tends to be lower due to the imbalanced class distribution.}
\label{tab:meta-eval}
\resizebox{\columnwidth}{!}{%
\setlength{\tabcolsep}{3pt} 
\begin{tabular}{ll c cccc cccc cccc}
\toprule
\multirow{2}{*}{\textbf{Dataset}} & \multirow{2}{*}{\textbf{Model}} & \multirow{2}{*}{\textbf{EM}} & \multicolumn{4}{c}{\textbf{Info-Think}} & \multicolumn{4}{c}{\textbf{Think-Search}} & \multicolumn{4}{c}{\textbf{Think-Ans}} \\
\cmidrule(lr){4-7} \cmidrule(lr){8-11} \cmidrule(lr){12-15}
& & & LLMaaJ & RM & Agree. & Kappa & LLMaaJ & NLI & Agree. & Kappa & SubEM & NLI & Agree. & Kappa \\
\midrule
\multirow{6}{*}{NQ} 
& SearchR1-7B-PPO & 0.480 & 0.685 & 0.712 & 0.880 & 0.689 & 0.935 & 0.951 & 0.947 & 0.058 & 0.808 & 0.583 & 0.756 & 0.456 \\
& SearchR1-7B-PPO-Format & 0.488 & 0.241 & 0.339 & 0.732 & 0.384 & 0.906 & 0.923 & 0.913 & 0.029 & 0.032 & 0.015 & 0.975 & 0.458 \\
& ReSearch-7B-Base & 0.396 & 0.707 & 0.720 & 0.878 & 0.645 & 0.869 & 0.885 & 0.933 & 0.241 & 0.877 & 0.836 & 0.873 & 0.484 \\
& ReSearch-7B-Instruct & 0.415 & 0.692 & 0.713 & 0.875 & 0.638 & 0.808 & 0.583 & 0.746 & 0.075 & 0.742 & 0.739 & 0.897 & 0.731 \\
& DeSA-3B & 0.375 & 0.182 & 0.177 & 0.962 & 0.881 & 0.705 & 0.734 & 0.910 & 0.773 & 0.452 & 0.726 & 0.464 & -0.027 \\
& DeSA-7B & 0.468 & 0.255 & 0.297 & 0.867 & 0.673 & 0.627 & 0.584 & 0.878 & 0.745 & 0.575 & 0.427 & 0.719 & 0.450 \\
\midrule
\multirow{6}{*}{HotpotQA} 
& SearchR1-7B-PPO & 0.433 & 0.564 & 0.575 & 0.847 & 0.666 & 0.817 & 0.767 & 0.831 & 0.133 & 0.774 & 0.461 & 0.680 & 0.385 \\
& SearchR1-7B-PPO-Format & 0.436 & 0.236 & 0.302 & 0.817 & 0.575 & 0.827 & 0.719 & 0.825 & 0.181 & 0.122 & 0.022 & 0.898 & 0.262 \\
& ReSearch-7B-Base & 0.406 & 0.586 & 0.565 & 0.833 & 0.607 & 0.808 & 0.752 & 0.872 & 0.311 & 0.912 & 0.813 & 0.874 & 0.479 \\
& ReSearch-7B-Instruct & 0.435 & 0.567 & 0.485 & 0.808 & 0.593 & 0.715 & 0.616 & 0.873 & 0.719 & 0.846 & 0.577 & 0.713 & 0.051 \\
& DeSA-3B & 0.352 & 0.081 & 0.085 & 0.974 & 0.846 & 0.361 & 0.636 & 0.488 & 0.048 & 0.688 & 0.764 & 0.838 & 0.586 \\
& DeSA-7B & 0.424 & 0.160 & 0.205 & 0.910 & 0.711 & 0.256 & 0.162 & 0.796 & 0.402 & 0.485 & 0.378 & 0.792 & 0.583 \\
\bottomrule
\end{tabular}%
}
\end{table}

\section{\ours: Training Faithful Search Agents with Structured Faithfulness Rewards}

\label{sec:improve_searchr1}
Our analysis in~\Cref{sec:evaluation} revealed a critical gap: RL agents trained with purely outcome-based rewards often fail to produce faithful reasoning, even when they arrive at the correct final answer. This finding highlights the need for direct supervision of intermediate reasoning quality. To fill this gap, we introduce \ours, a framework designed to train search agents that are not only correct but also verifiably faithful. An overview of the pipeline with PPO RL algorithm is shown in~\cref{fig:ours-pipeline}. The core principle of \ours is to enrich the RL reward signal with fine-grained feedback at a small, fixed set of structurally meaningful transitions in the agentic search trajectory\,---\,a checklist-style alternative to per-step process reward models. This is achieved through two key components: a multi-faceted reward function that balances task accuracy with reasoning faithfulness, and a practical, distilled reward model to efficiently supervise the RL training.

\noindent\textbf{Reward design.}
The foundation of \ours is a reward function that combines the conventional outcome-based reward with our proposed structured faithfulness rewards. Denote the outcome-based reward as $\mathcal{R}_{\textit{EM}}$, measured by the exact match between the predicted answer and the groundtruth as in \searchrone. For the two key faithfulness dimensions we optimize during training, we define their corresponding rewards as $\mathcal{R}_{\textit{info-think}}$ and $\mathcal{R}_{\textit{think-answer}}$. The final reward function is a weighted sum:
\begin{equation}
    \mathcal{R} = w_{\textit{EM}} \, \cdot \, \mathcal{R}_{\textit{EM}} + w_{\textit{info-think}} \, \cdot \, \mathcal{R}_{\textit{info-think}} + w_{\textit{think-search}} \, \cdot \, \mathcal{R}_{\textit{think-search}} + w_{\textit{think-answer}} \, \cdot \, \mathcal{R}_{\textit{think-answer}},
    \label{eq:reward-design}
\end{equation}
where $w$ are hyperparameters representing the weight of each corresponding reward component. This reward formulation explicitly optimizes the agent to balance task accuracy (EM) and the faithfulness of its reasoning process. We ground our implementation of \ours on the \searchrone framework. 

Practically, we opted to set $w_{\textit{think-search}}=0$, i.e., skipping the Think-Search term. The reason of this choice is three-fold: (i) quantitatively, we observed in~\cref{sec:evaluation} that all checkpoints we evaluated are already showing high Think-Search faithfulness; (ii) in our manual inspections of the \searchrone trajectories, we found that the model has developed a consistent pattern of ``first reason about its information need, then write the search query'' (case studies in~\cref{asec:case_studies}), suggesting that training with EM reward already equipped the policy model with robust think-search faithfulness; (iii)
training search agents is time-consuming and computationally expensive. Removing this term eliminates an extra hyperparameter, thereby streamlining the training process and enhancing the efficiency.

We note that \ours can be considered a conceptual framework that incorporates faithfulness as a structured reward signal to improve the search agent. The exact instantiation of each individual reward can be implemented differently. In this work, we use a model-based approach for information-think faithfulness and substring exact-match for think-answer faithfulness, and leave more exploration of other reward implementations --- such as rich semantic matching signals for think-answer faithfulness --- to future work.

\noindent\textbf{Practical implementation via a distilled reward model.}
\label{subsec:reward_modeling}
The cost and latency of LLMaaJ approach make it impractical for scalable on-policy RL training. To tackle this challenge, a core component of the \ours framework is a smaller, distilled reward model (RM) trained to replicate the judgments of the larger LLMaaJ. We focus exclusively on information-think faithfulness.

To train the RM, we first collect a large-scale dataset of \searchrone trajectories. We subsample 27,000 samples from a combination of the NQ~\citep{kwiatkowski2019nqdata} and HotpotQA~\citep{yang2018hotpotqa} training splits. We prompt Claude-3.7-Sonnet to label these instances for Information-Think faithfulness (see Appendix for prompt templates). We then split the 27K instances into 24K for training and 3K for evaluation. Using LoRA~\citep{hu2022lora}, we fine-tune a Qwen2.5-14B-Instruct model, which has high consistency (0.899) and a strong Cohen's $\kappa$ coefficient (0.797) with the Claude-3.7-Sonnet labels. To validate our RM, an author of this paper manually annotated 100 examples from the test set. Both Claude-3.7-Sonnet and our trained RM show high consistency with human judgments (0.850 and 0.910, respectively, with corresponding Cohen's $\kappa$ of 0.699 and 0.819; see~\cref{tab:human_eval}). Given its strong performance and efficiency, we use this fine-tuned RM for all subsequent RL training and evaluation.

Further, note that the datasets for Info-Think RM are labeled by Claude-3.7-Sonnet rather than Claude-4.5-Sonnet which is the judge model in~\cref{subsec:evaluating_searchr1_faithfulness}. There are two main reasons: (i) The API of Claude-3.7-Sonnet becomes deprecated when we are extending the faithfulness evaluation to more agentic search models. 
Re-labeling the large datasets using Claude-4.5-Sonnet and retraining the RM on the newly labeled data can be costly. (ii) Our evaluation results in~\Cref{tab:rm_info_thinkv2} show that Claude-3.7-Sonnet and Claude-4.5-Sonnet have high consistency ratio and $\kappa$ coefficient, meaning that the two models are highly consistent. Therefore, we still use the datasets labeled by Claude-3.7-Sonnet for RM's training and evaluation.
Further training details and evaluation results, including the human evaluation analysis are in~\cref{asec:reward_model_scaling}.

\noindent\textbf{Training curriculum.}
In preliminary RL experiments, we found that directly applying the faithfulness rewards $\mathcal{R}_{\textit{info-think}}$ and $\mathcal{R}_{\textit{think-ans}}$ substantially improved faithfulness metrics but slightly reduced task performance compared to \searchrone. We hypothesize that introducing these rewards too early restricts the policy model’s exploration of diverse reasoning strategies, causing it to over-optimize for faithfulness --- a form of reward hacking. To mitigate this, we adopt a curriculum learning-inspired strategy: from step 1 to $T_1$, we use only outcome-based rewards by setting $w_{\textit{info-think}}$ and $w_{\textit{think-ans}}$ in~\cref{eq:reward-design} to 0; from $T_1+1$ to $T_2$, we linearly warm up these weights (note that this reward warmup is separate from the linear learning rate schedule); and from $T_2$ onward, we apply the full rewards. Practically, $T_1$ is set to the end of the first training epoch, and the warmup lasts 0.5 epoch.

To summarize, the \ours framework provides a practical methodology for structured supervision of intermediate reasoning by combining a multi-faceted reward function with an efficient, distilled reward model. This enables direct optimization of reasoning faithfulness during RL training without sacrificing performance or scalability.

\begin{table*}[t!]
\centering
\caption{Performance of different methods on selected QA datasets. $^\dagger$ denotes in-domain datasets and $^*$ denotes out-of-domain dataset. All the baseline results are from~\citet{jin2025searchr1}. \ours-R1 is \colorbox{green!10}{\searchrone} trained with our \ours framework. $^\clubsuit$ indicates statistically significant compared to Search-R1-7B-Base-PPO, while $^\spadesuit$ indicates significant compared to Search-R1-7B-Base-PPO w/ Format.}
\label{tab:main_performance_em}
\resizebox{\textwidth}{!}{
\begin{tabular}{@{}lcccccccr@{}}
\toprule
\multirow{2}{*}{\textbf{Methods}} & \multicolumn{3}{c}{\textbf{General QA}} & \multicolumn{4}{c}{\textbf{Multi-Hop QA}} & \multirow{2}{*}{\textbf{Avg.}} \\
\cmidrule(r){2-4} \cmidrule(lr){5-8}
& NQ$^\dagger$ & TriviaQA$^*$ & PopQA$^*$ & HotpotQA$^\dagger$ & 2wiki$^*$ & MuSiQue$^*$ & Bamboogle$^*$ & \\
\midrule
Direct Inference & 0.134 & 0.408 & 0.140 & 0.183 & 0.250 & 0.031 & 0.120 & 0.181 \\
CoT & 0.048 & 0.185 & 0.054 & 0.092 & 0.111 & 0.022 & 0.232 & 0.106 \\
IRCoT & 0.224 & 0.478 & 0.301 & 0.133 & 0.149 & 0.072 & 0.224 & 0.239 \\
Search-o1 & 0.151 & 0.443 & 0.131 & 0.187 & 0.176 & 0.058 & 0.296 & 0.206 \\
RAG & 0.349 & 0.585 & 0.392 & 0.299 & 0.235 & 0.058 & 0.208 & 0.304 \\
SFT & 0.318 & 0.354 & 0.121 & 0.217 & 0.259 & 0.066 & 0.112 & 0.207 \\
R1-base & 0.297 & 0.539 & 0.202 & 0.242 & 0.273 & 0.083 & 0.296 & 0.276 \\
R1-instruct & 0.270 & 0.537 & 0.199 & 0.237 & 0.292 & 0.072 & 0.293 & 0.271 \\
Rejection Sampling & 0.360 & 0.592 & 0.380 & 0.331 & 0.296 & 0.123 & 0.355 & 0.348 \\ \midrule
\desa-3B-Instruct-GRPO & 0.375 & 0.575 & 0.397 & 0.352 & 0.363 & 0.134 & 0.347 & 0.363 \\
\desa-7B-Instrct-GRPO & 0.468 & 0.631 & 0.440 & 0.424 & 0.374 & 0.197 & 0.395 & 0.418 \\
\research-7B-Base & 0.396 & 0.606 & 0.442 & 0.406 & \underline{0.447} & \underline{0.217} & 0.432 & 0.422 \\
\research-7B-Instruct & 0.415 & 0.640 & 0.450 & 0.435 & \textbf{0.476} & \textbf{0.223} & 0.424 & 0.438 \\
Search-R1-7B-Base-PPO w/ Format & \textbf{0.488} & 0.644 & \textbf{0.469} & 0.436  & 0.412 & 0.187 & 0.403 & 0.434 \\
\rowcolor{green!10} 
Search-R1-7B-Base-PPO & 0.480 & 0.638 & 0.457 & 0.433 & 0.382 & 0.196 & 0.432 & 0.431 \\
\midrule

\multicolumn{9}{l}{\textit{\textsc{VERITAS-R1}}} \\
\rowcolor{green!10} 
EM+Info-Think & \underline{0.486} & \underline{0.650} & 0.463 & \underline{0.445}  & 0.423 & 0.206 & \underline{0.456} & \textbf{0.447}$^{\clubsuit\spadesuit}$ \\
\rowcolor{green!10} 
EM+Think-Ans & 0.482 & \textbf{0.658} & 0.464 & \underline{0.445}  & 0.420 & 0.189 & 0.416 & 0.439 \\
\rowcolor{green!10} 
EM+Info-Think+Think-Ans & 0.484 & 0.645 & \underline{0.466} & \textbf{0.446}  & 0.419 & 0.192 & \textbf{0.464} & \underline{0.445}$^\clubsuit$ \\

\bottomrule
\end{tabular}
}
\label{tab:main_em}
\end{table*}

\section{Experiments}
\label{sec:results}
We conduct a series of experiments to validate our central hypothesis: that incorporating structured faithfulness rewards via the \ours framework can improve both the reliability of an agent's reasoning and its final task performance. We focus on the following key research questions:
\begin{itemize}[]
\item[\textbf{RQ1:}] Does training with \ours lead to improvement in reasoning faithfulness?
\item[\textbf{RQ2:}] How do these structured rewards affect final task accuracy compared to purely outcome-based RL training?
\end{itemize}

\subsection{Experiment Setup}
\label{subsec:experimental_setup}

\noindent
\textbf{Models and datasets.} To ensure a fair comparison, we build our implementation directly upon the \searchrone framework. We use the Qwen2.5-7B-Base model~\citep{yang2024qwen25} as our policy model and PPO~\citep{schulman2017proximal} as the RL algorithm, mirroring the best-performing configuration of the original \searchrone. Evaluation spans seven diverse QA benchmarks: NQ~\citep{kwiatkowski2019nqdata}, TriviaQA~\citep{joshi2017triviaqa}, PopQA~\citep{mallen-etal-2023-whennottotrust}, HotpotQA~\citep{yang2018hotpotqa}, 2Wiki~\citep{ho-etal-2020-constructing}, MuSiQue~\citep{trivedi2022musique}, and Bamboogle~\citep{press2023bamboogle}. These datasets cover a range of complexities from general to multi-hop QA, allowing for a comprehensive assessment.

\noindent
\textbf{Baselines.} We compare our model, which we refer to as \textbf{\ours-R1} (i.e., \searchrone trained with the \ours framework), against a comprehensive set of baselines. Our primary comparison is against the original \searchrone~\citep{jin2025searchr1}, which uses outcome-only RL. We also include other baselines, detailed in~\cref{asec:baselines}.

\noindent
\textbf{Metrics.} To answer our research questions, we evaluate all models on: (i) \textit{Task Performance}, measured by Exact Match (EM), and (ii) \textit{Reasoning Faithfulness}, using our proposed Information-Think and Think-Answer faithfulness metrics (\cref{subsec:evaluating_faithfulness}). For task performance, we report dataset-level paired Wilcoxon signed-rank tests across the seven benchmarks ($n=7$) at the $p=0.05$ level, following standard practice for comparing learned models over multiple datasets~\citep{demvsar2006statistical,dror-etal-2018-hitchhikers}. We adopt this dataset-level paired protocol rather than aggregating across training seeds because (i) RL training of search agents is computationally intensive\,---\,each run requires tens of high-memory GPUs over several days\,---\,making large seed sweeps infeasible, and (ii) hypothesis tests over a small number of seeds are known to have very low statistical power and produce unreliable conclusions~\citep{henderson-etal-2018-deepreinforcement}. As a stability check, we additionally re-ran the Info-Think variant of \ours-R1 with three independent seeds and observed average EM of $\{0.442, 0.446, 0.445\}$ (std $0.0021$), consistent with the result reported in~\cref{tab:main_performance_em}.

\noindent
\textbf{Implementation details.} Our reward model is a LoRA-finetuned Qwen2.5-14B-Instruct~\citep{yang2024qwen25}, trained as described in~\cref{subsec:reward_modeling} using LLaMA-Factory~\citep{zheng2024llamafactory}. For the \ours-R1 policy model, we tune the reward weights and find an optimal balance with $w_{\textit{EM}}=1.0$, $w_{\textit{info-think}}=0.05$, and $w_{\textit{think-ans}}=0.02$ (see~\cref{fig:hyperparam-sensitivity} for an ablation). Other training parameters, such as learning rate (1e-6) and batch size (256), as well as the retriever (E5-base-v2~\citep{wang2022text}) and corpus (2018 Wikipedia dump), are kept identical to the original \searchrone setup for a controlled comparison. All experiments were conducted on 32x NVIDIA A100 GPUs.

\begin{figure*}[t]
    \centering
    \includegraphics[width=\textwidth]{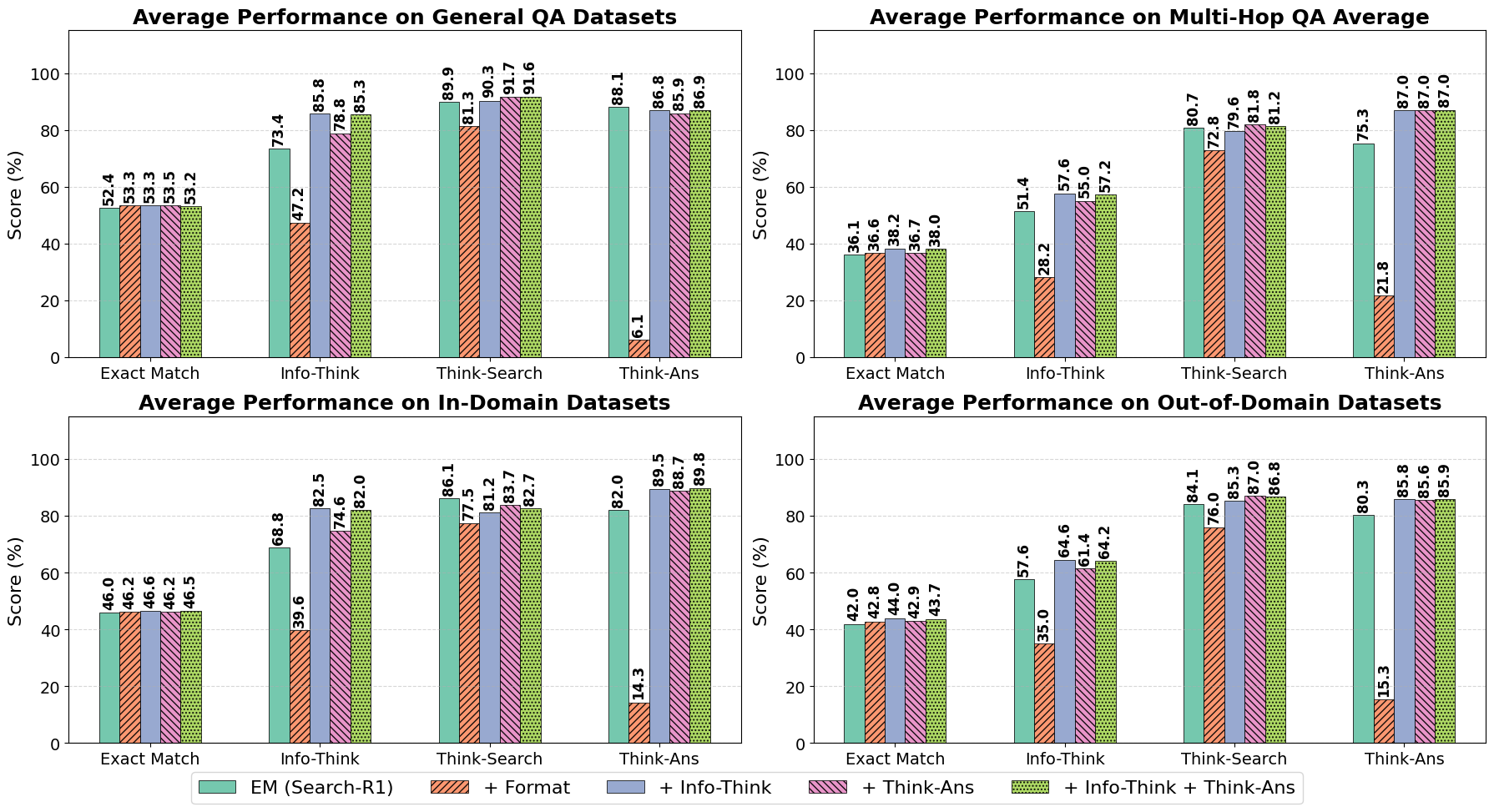}
    \caption{Faithfulness evaluation results comparing two variants of Search-R1 and our methods. }
    \label{fig:search_r1_faithfulness_main}
\end{figure*}

\subsection{Results and Analysis}
\label{subsec:results-analysis}

\begin{table*}[t]
\centering
\caption{Model Performance of the proposed Faithful Search-R1. (EM results are from our machines and may differ slightly from~\cref{tab:main_performance_em} due to vLLM randomness).}
\label{tab:main_performance_faithfulness}
\resizebox{\columnwidth}{!}{
\begin{tabular}{@{}l ccccccc c@{}}
\toprule
\textbf{Method} & \textbf{NQ} & \textbf{TriviaQA} & \textbf{PopQA} & \textbf{HotpotQA} & \textbf{2wiki} & \textbf{MuSiQue} & \textbf{Bamboogle} & \textbf{Average} \\
\midrule
\multicolumn{9}{c}{\textit{Exact Match (EM)}} \\
\midrule
Search-R1 (Baseline) & 0.482 & 0.641 & 0.450 & 0.437 & 0.381 & 0.201 & 0.425 & 0.431 \\
+ Format (Baseline) & 0.494 & 0.643 & 0.462 & 0.430 & 0.413 & 0.196 & 0.425 & 0.436 \\
\ours Info-Think         & 0.486 & 0.650 & 0.463 & 0.445 & 0.423 & 0.206 & 0.456 & 0.447 \\
\ours Think-Ans          & 0.482 & 0.658 & 0.464 & 0.442 & 0.420 & 0.189 & 0.416 & 0.439 \\
\ours Info-Think+Think-Ans & 0.484 & 0.645 & 0.466 & 0.446 & 0.419 & 0.192 & 0.464 & 0.445 \\
\midrule
\multicolumn{9}{c}{\textit{Info-Think Faithfulness}} \\
\midrule
Search-R1 (Baseline) & 0.762 & 0.745 & 0.695 & 0.614 & 0.349 & 0.442 & 0.651 & 0.608 \\
+ Format (Baseline)            & 0.429 & 0.534 & 0.452 & 0.364 & 0.190 & 0.256 & 0.317 & 0.363 \\
\ours Info-Think         & 0.933 & 0.821 & 0.819 & 0.718 & 0.354 & 0.490 & 0.744 & 0.697 \\
\ours Think-Ans          & 0.795 & 0.778 & 0.792 & 0.698 & 0.332 & 0.452 & 0.717 & 0.652 \\
\ours Info-Think + Think-Ans & 0.929 & 0.811 & 0.820 & 0.712 & 0.356 & 0.489 & 0.732 & 0.693 \\
\midrule
\multicolumn{9}{c}{\textit{Think-Search Faithfulness}} \\
\midrule
Search-R1 (Baseline) & 0.950 & 0.867 & 0.880 & 0.771 & 0.744 & 0.846 & 0.867 & 0.846 \\
+ Format (Baseline) & 0.857 & 0.774 & 0.808 & 0.693 & 0.739 & 0.676 & 0.802 & 0.764 \\
\ours Info-Think & 0.855 & 0.919 & 0.934 & 0.769 & 0.745 & 0.848 & 0.821 & 0.842 \\
\ours Think-Ans & 0.882 & 0.933 & 0.935 & 0.792 & 0.769 & 0.855 & 0.857 & 0.860 \\
\ours Info-Think+Think-Ans & 0.879 & 0.934 & 0.934 & 0.775 & 0.774 & 0.851 & 0.848 & 0.856 \\

\midrule
\multicolumn{9}{c}{\textit{Think-Ans Faithfulness}} \\
\midrule
Search-R1 (Baseline) & 0.836 & 0.898 & 0.909 & 0.803 & 0.644 & 0.670 & 0.895 & 0.808 \\
+ Format (Baseline)       & 0.052 & 0.064 & 0.066 & 0.235 & 0.327 & 0.195 & 0.113 & 0.150 \\
\ours Info-Think         & 0.892 & 0.844 & 0.868 & 0.899 & 0.872 & 0.819 & 0.889 & 0.869 \\
\ours Think-Ans          & 0.879 & 0.855 & 0.842 & 0.895 & 0.869 & 0.838 & 0.877 & 0.865 \\
\ours Info-Think+Think-Ans & 0.893 & 0.838 & 0.877 & 0.903 & 0.867 & 0.823 & 0.888 & 0.870 \\

\bottomrule
\end{tabular}
}
\end{table*}

Our results demonstrate that \ours successfully improves both faithfulness and task accuracy, confirming our central hypothesis. We find that explicitly rewarding the reasoning process creates a synergistic effect, leading to more robust and effective agents.

\noindent
\textbf{\ours improves reasoning faithfulness (RQ1).}
As shown in~\cref{fig:search_r1_faithfulness_main}, training with \ours rewards leads to gains in faithfulness. For example, incorporating the $\mathcal{R}_{\textit{info-think}}$ reward can largely improves Information-Think faithfulness across all dataset categories. For example, on general QA datasets, average Information-Think faithfulness increases from 0.734 for the baseline \searchrone to 0.853 for \ours-R1. Interestingly, \ours-R1 further enhances \searchrone's Think-Search Faithfulness on both general QA and Multi-Hop QA datasets, even though we do not adopt think-search faithfulness as the training signal. This directly validates that our structured faithfulness rewards successfully steer the agent towards generating more grounded and verifiable reasoning steps. Gains in think--answer faithfulness are clear on multi-hop datasets. However, the improvement of think--answer faithfulness is not stable on general QA datasets.

\noindent
\textbf{Improved faithfulness translates to higher accuracy (RQ2).}
Crucially, the improvement in reasoning quality translates to better final performance. As detailed in~\cref{tab:main_performance_em}, \ours-R1 consistently outperforms the strong \searchrone baseline across the majority of datasets. On the challenging multi-hop datasets, \ours-R1 improves the average EM score from 0.361 to 0.380. The average EM score on general QA datasets is also increased from 0.524 to 0.532. This demonstrates that encouraging the model to "think" more faithfully is not a constraint but a facilitator for finding the correct answer, effectively reducing reward hacking and promoting more robust problem-solving strategies. The detailed performance on all seven datasets can be found at~\cref{tab:main_performance_faithfulness}.
\noindent

\subsection{Ablation Studies}
\noindent
\textbf{Analysis of reward components.}
Our analysis also reveals nuances in how different faithfulness rewards affect behavior. While the Information-Think reward generally provides a positive signal for faithfulness and EM, the Think-Answer reward has a more complex effect. As seen in~\cref{fig:search_r1_faithfulness_main}, although adding $\mathcal{R}_{\textit{think-answer}}$ improves its corresponding metric on a macro average (0.865 vs 0.808 averaged over 7 datasets), it leads to a slight drop on TriviaQA and PopQA datasets. This suggests that ensuring thoughts are grounded in evidence ($\mathcal{R}_{\textit{info-think}}$) is a more effective and stable mechanism for improving overall search model quality than enforcing a strict logical entailment to the final answer during training.

\noindent\textbf{Hyperparameter sensitivity.}
\label{subsec:hyperparameter_sensitivity}
\begin{figure}[t]
    \centering
    \includegraphics[width=\columnwidth]{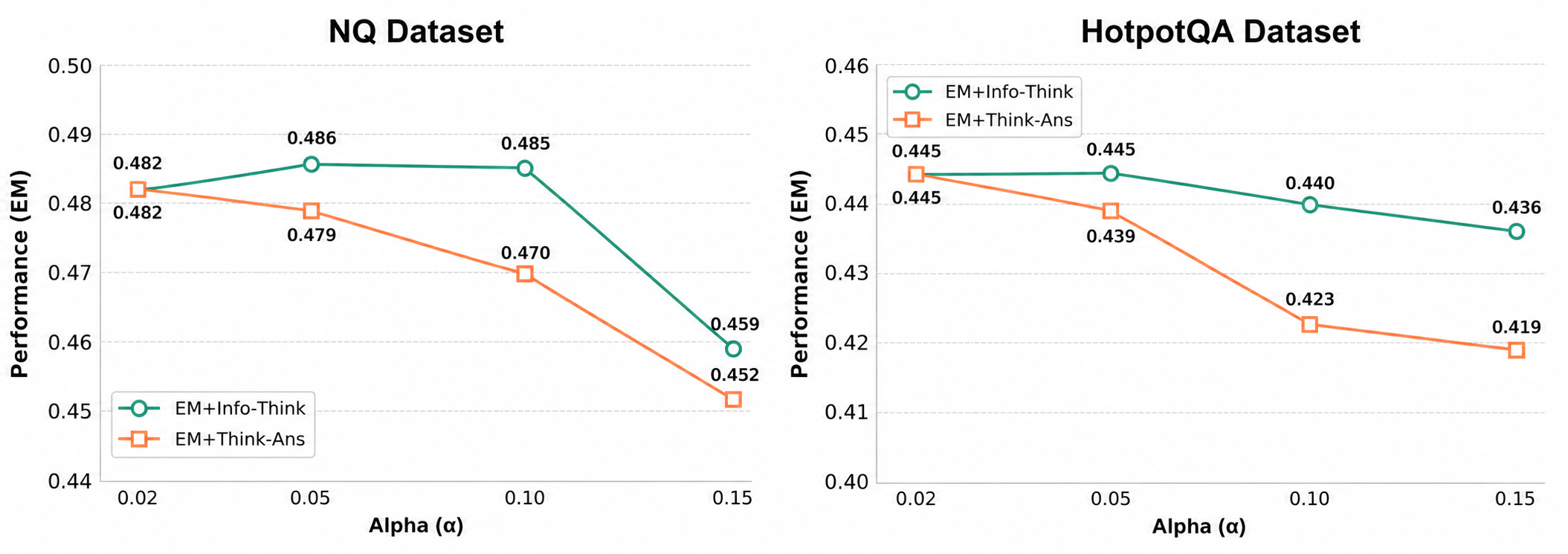}
    \caption{Hyperparameter sensitivity study.}
    \label{fig:hyperparam-sensitivity}
\end{figure}
We experiment with different hyperparameters $w_{\textit{info-think}},w_{\textit{think-ans}}$ and report results in~\cref{fig:hyperparam-sensitivity}. We notice that increasing both hyperparameters to greater than $0.05$ leads to performance degradation, as the policy model overly focus on the faithfulness reward. Therefore, we opted to use $w_{\textit{info-think}}=0.05$, $w_{\textit{think-ans}}=0.02$ in our final combined reward (EM+Info-Think+Think-Ans).

\begin{table*}[t]
\centering
\caption{Effects of the training curriculum to task performance. We use the \ours Info-Think variant to show the tradeoff between task performance and Info-Think faithfulness between two training strategies.}
\label{tab:ablation_curriculum}
\resizebox{0.9\columnwidth}{!}{
\begin{tabular}{@{}l ccccccc c@{}}
\toprule
\textbf{Method} & \textbf{NQ} & \textbf{TriviaQA} & \textbf{PopQA} & \textbf{HotpotQA} & \textbf{2wiki} & \textbf{MuSiQue} & \textbf{Bamboogle} & \textbf{Average} \\
\midrule
\multicolumn{9}{l}{\textit{Exact Match (EM)}} \\
w/o curriculum & 0.471 & 0.641 & 0.480 & 0.437 & 0.429 & 0.174 & 0.400 & 0.433 \\
w/ curriculum & 0.486 & 0.650 & 0.463 & 0.445 & 0.423 & 0.206 & 0.456 & 0.447 \\
\midrule
\multicolumn{9}{l}{\textit{Info-Think Faithfulness}} \\
w/o curriculum& 0.956 & 0.832 & 0.826 & 0.786 & 0.595 & 0.619 & 0.749 & 0.766 \\
w/ curriculum & 0.933 & 0.821 & 0.819 & 0.718 & 0.354 & 0.490 & 0.744 & 0.697 \\
\bottomrule
\end{tabular}
}
\end{table*}

\noindent\textbf{Effects of the training curriculum.}
We show an ablation study of the training curriculum in~\cref{tab:ablation_curriculum}. 
We use a simplified setting to only compare two reward signals --- EM and Info-Think faithfulness --- and their corresponding evaluation results. We notice that directly adding the faithfulness reward at the start of the training makes the policy model learn to hack the faithfulness reward, leading the lower EM results at the end of the training. The training dynamics (\cref{fig:training_dynamics}) also suggest that directly adding the faithfulness reward at the start of the training limits the policy model's exploration.

\subsection{Discussions}
\label{subsec:discussion}
\noindent\textbf{Limitation and improvement of model-based evaluations.}
Although using LLMs to evaluate results has been a common practice, it has been reported in the literature that they still have several limitations~\citep{li2024llmsasjudgescomprehensivesurvey}, such as position bias~\citep{shi2025llmajpositionbias}, token/label bias~\citep{jiang-etal-2024-peek,xu-etal-2024-context} and overconfidence~\citep{khan2024debatingwithmorepersuasive}. Recent works have proposed corresponding mitigations, such as calibration~\citep{wang-etal-2024-large-language-models-fair,lee2026correctlyreportllmasajudgeevaluations} or aggregating predictions from multiple LLMs~\citep{verga2024replacingjudgesjuriesevaluating,zhao2025frommanyvoicestoone}.

In this work, we opted for a uni-dimensional scalar evaluation for each individual faithfulness dimension. Recent works have also explored multi-dimensional, structured evaluation criteria, commonly referred to as ``\emph{checklists}'' or ``\emph{rubrics}''~\citep{ribeiro-etal-2020-beyond,arora2025healthbenchevaluatinglargelanguage,deshpande-etal-2025-multichallenge}, and further incorporate this fine-grained signals into RL training~\citep[\emph{inter alia}]{gunjal2026rubricsasrewards,viswanathan2025checklists,shao2025drtulureinforcementlearning}. Due to the limited bandwidth, we leave a more comprehensive investigation of faithfulness evaluation and corresponding RL training to the future work.

We also note that our evaluation pipeline is explicitly designed to mitigate circularity between the training and evaluation signals. The held-out judge used throughout~\cref{sec:evaluation} (Claude-4.5-Sonnet) differs from the labeling source of our distilled reward model (Claude-3.7-Sonnet, see~\cref{subsec:reward_modeling}), so the faithfulness numbers reported in~\cref{sec:evaluation} are decoupled from the training reward. We additionally validate the distilled reward model against human annotations (\cref{tab:human_eval}), where it achieves $0.910$ consistency and Cohen's $\kappa{=}0.819$ over 100 samples\,---\,higher than either Claude-3.7-Sonnet or Claude-4.5-Sonnet on the same set\,---\,suggesting it serves as a reliable proxy rather than introducing arbitrary bias.

\section{Conclusion and Future Works}
\label{sec:conclusion}
In this work, we addressed the issue of unfaithful reasoning in RL-based agentic search, where models optimized for final answers often produce untrustworthy intermediate steps. We introduced a formal evaluation framework with three faithfulness metrics and proposed \ours, a training paradigm that integrates these metrics as structured, checklist-style rewards. Our central finding is that rewarding the reasoning process does not trade off with performance; instead, it creates a positive synergy across diverse QA benchmarks, leading to agents that are not only more faithful but also achieve higher task accuracy. This work underscores the value of structured supervision over intermediate reasoning for developing more reliable and transparent AI agents.

More broadly, \ours can be viewed as decomposing a holistic notion of trajectory quality\,---\,faithfulness\,---\,into a small set of structured, rubric-like sub-criteria (think--search, information--think, think--answer), each instantiated as an independent reward signal. This connects naturally to the emerging line of rubric- and checklist-based evaluation~\citep{ribeiro-etal-2020-beyond,arora2025healthbenchevaluatinglargelanguage,deshpande-etal-2025-multichallenge} and rubric-as-reward RL~\citep[\emph{inter alia}]{gunjal2026rubricsasrewards,viswanathan2025checklists,shao2025drtulureinforcementlearning}, which argue that complex, multi-aspect objectives are better learned from structured criteria than from a single scalar judge. Our results suggest that even a coarse three-dimensional rubric is sufficient to recover meaningful gains in both faithfulness and accuracy. We view this as evidence for a broader principle: as outcome-only rewards saturate, the next axis of improvement for agentic systems may lie in spending more compute on \emph{evaluation}\,---\,via richer, finer-grained, possibly automatically-generated rubrics\,---\,and propagating those signals back into training. Future work could focus on automatically generating query-conditioned faithfulness rubrics, scalable methods for distilling such fine-grained rewards, and extending this paradigm beyond agentic search to other long-horizon reasoning tasks.

\section*{Limitations and Potential Risks}
\label{sec:limitation}
While our work demonstrates the benefits of incorporating faithfulness rewards, we acknowledge that there are still several limitations in our work. First, our evaluation of Information-Think Faithfulness relies on an LLM-as-a-Judge, which we subsequently use our trained reward model. Although we show this approach is effective, it is inherently subject to the biases and potential errors of the judge model. It is worth exploring the development of more objective, non–model-based metrics for evaluating faithfulness in agentic search models.
Second, our Think-Answer Faithfulness metric is based on substring exact-match (Sub-EM), which is precise but may lack recall. It can fail to recognize legitimate paraphrasing or logical inference, potentially treating a faithful answer as unfaithful. More sophisticated semantic matching techniques could provide a more robust evaluation.
Lastly, our experiments were conducted on open-domain question-answering tasks. The effectiveness of our proposed rewards may vary in other domains, such as enterprise search or medical QA, where the nature of evidence and reasoning can be substantially different. Future work should explore the generalizability of these faithfulness-aware training methods to a wider range of applications.

We use public benchmarks licensed for academic usage. Our small scale human evaluation is conducted by an author of this paper, who is a trained NLP researcher. To the best of our knowledge, this paper does not incur potential risks and ethical concerns.

\bibliography{tmlr-style-file-main/main}

\begin{thebibliography}{123}
\providecommand{\natexlab}[1]{#1}
\providecommand{\url}[1]{\texttt{#1}}
\expandafter\ifx\csname urlstyle\endcsname\relax
  \providecommand{\doi}[1]{doi: #1}\else
  \providecommand{\doi}{doi: \begingroup \urlstyle{rm}\Url}\fi

\bibitem[Ahmadian et~al.(2024)Ahmadian, Cremer, Gall{\'e}, Fadaee, Kreutzer, Pietquin, {\"U}st{\"u}n, and Hooker]{ahmadian-etal-2024-back}
Arash Ahmadian, Chris Cremer, Matthias Gall{\'e}, Marzieh Fadaee, Julia Kreutzer, Olivier Pietquin, Ahmet {\"U}st{\"u}n, and Sara Hooker.
\newblock Back to basics: Revisiting {REINFORCE}-style optimization for learning from human feedback in {LLM}s.
\newblock In Lun-Wei Ku, Andre Martins, and Vivek Srikumar (eds.), \emph{Proceedings of the 62nd Annual Meeting of the Association for Computational Linguistics (Volume 1: Long Papers)}, pp.\  12248--12267, Bangkok, Thailand, August 2024. Association for Computational Linguistics.
\newblock \doi{10.18653/v1/2024.acl-long.662}.
\newblock URL \url{https://aclanthology.org/2024.acl-long.662/}.

\bibitem[Ahn et~al.(2024)Ahn, Verma, Lou, Liu, Zhang, and Yin]{ahn2024rejectsample}
Janice Ahn, Rishu Verma, Renze Lou, Di~Liu, Rui Zhang, and Wenpeng Yin.
\newblock Large language models for mathematical reasoning: Progresses and challenges.
\newblock In Neele Falk, Sara Papi, and Mike Zhang (eds.), \emph{Proceedings of the 18th Conference of the European Chapter of the Association for Computational Linguistics: Student Research Workshop}, pp.\  225--237, St. Julian{'}s, Malta, March 2024. Association for Computational Linguistics.
\newblock \doi{10.18653/v1/2024.eacl-srw.17}.
\newblock URL \url{https://aclanthology.org/2024.eacl-srw.17/}.

\bibitem[{Anthropic}(2025)]{anthropic_claude_3_7}
{Anthropic}.
\newblock Claude 3.7 sonnet and claude code, February 2025.

\bibitem[Arcuschin et~al.(2025)Arcuschin, Janiak, Krzyzanowski, Rajamanoharan, Nanda, and Conmy]{arcuschin2025cotfaitheval}
Iván Arcuschin, Jett Janiak, Robert Krzyzanowski, Senthooran Rajamanoharan, Neel Nanda, and Arthur Conmy.
\newblock Chain-of-thought reasoning in the wild is not always faithful, 2025.
\newblock URL \url{https://arxiv.org/abs/2503.08679}.

\bibitem[Arora et~al.(2025)Arora, Wei, Hicks, Bowman, Quiñonero-Candela, Tsimpourlas, Sharman, Shah, Vallone, Beutel, Heidecke, and Singhal]{arora2025healthbenchevaluatinglargelanguage}
Rahul~K. Arora, Jason Wei, Rebecca~Soskin Hicks, Preston Bowman, Joaquin Quiñonero-Candela, Foivos Tsimpourlas, Michael Sharman, Meghan Shah, Andrea Vallone, Alex Beutel, Johannes Heidecke, and Karan Singhal.
\newblock Healthbench: Evaluating large language models towards improved human health, 2025.
\newblock URL \url{https://arxiv.org/abs/2505.08775}.

\bibitem[Asai et~al.(2024)Asai, Wu, Wang, Sil, and Hajishirzi]{asai2023selfrag}
Akari Asai, Zeqiu Wu, Yizhong Wang, Avirup Sil, and Hannaneh Hajishirzi.
\newblock Self-{RAG}: Learning to retrieve, generate, and critique through self-reflection.
\newblock In \emph{The Twelfth International Conference on Learning Representations}, 2024.
\newblock URL \url{https://openreview.net/forum?id=hSyW5go0v8}.

\bibitem[Atanasova et~al.(2023)Atanasova, Camburu, Lioma, Lukasiewicz, Simonsen, and Augenstein]{atanasova2023counterfactualfaithfulness}
Pepa Atanasova, Oana-Maria Camburu, Christina Lioma, Thomas Lukasiewicz, Jakob~Grue Simonsen, and Isabelle Augenstein.
\newblock Faithfulness tests for natural language explanations.
\newblock In Anna Rogers, Jordan Boyd-Graber, and Naoaki Okazaki (eds.), \emph{Proceedings of the 61st Annual Meeting of the Association for Computational Linguistics (Volume 2: Short Papers)}, pp.\  283--294, Toronto, Canada, July 2023. Association for Computational Linguistics.
\newblock \doi{10.18653/v1/2023.acl-short.25}.
\newblock URL \url{https://aclanthology.org/2023.acl-short.25/}.

\bibitem[Baker et~al.(2025)Baker, Huizinga, Gao, Dou, Guan, Madry, Zaremba, Pachocki, and Farhi]{baker2025monitoring}
Bowen Baker, Joost Huizinga, Leo Gao, Zehao Dou, Melody~Y. Guan, Aleksander Madry, Wojciech Zaremba, Jakub Pachocki, and David Farhi.
\newblock Monitoring reasoning models for misbehavior and the risks of promoting obfuscation, 2025.
\newblock URL \url{https://arxiv.org/abs/2503.11926}.

\bibitem[Balasubramanian et~al.(2025)Balasubramanian, Basu, and Feizi]{balasubramanian2025cotbias}
Sriram Balasubramanian, Samyadeep Basu, and Soheil Feizi.
\newblock A closer look at bias and chain-of-thought faithfulness of large (vision) language models.
\newblock In Christos Christodoulopoulos, Tanmoy Chakraborty, Carolyn Rose, and Violet Peng (eds.), \emph{Findings of the Association for Computational Linguistics: EMNLP 2025}, pp.\  13406--13439, Suzhou, China, November 2025. Association for Computational Linguistics.
\newblock ISBN 979-8-89176-335-7.
\newblock \doi{10.18653/v1/2025.findings-emnlp.723}.
\newblock URL \url{https://aclanthology.org/2025.findings-emnlp.723/}.

\bibitem[Bao et~al.(2025)Bao, Zhang, Wang, Yang, and Zhang]{bao-etal-2025-howlikely}
Guangsheng Bao, Hongbo Zhang, Cunxiang Wang, Linyi Yang, and Yue Zhang.
\newblock How likely do {LLM}s with {C}o{T} mimic human reasoning?
\newblock In \emph{Proceedings of the 31st International Conference on Computational Linguistics}, pp.\  7831--7850, 2025.

\bibitem[Bentham et~al.(2024)Bentham, Stringham, and Marasovic]{bentham2024chainofthoughtunfaithfulness}
Oliver Bentham, Nathan Stringham, and Ana Marasovic.
\newblock Chain-of-thought unfaithfulness as disguised accuracy.
\newblock \emph{Transactions on Machine Learning Research}, 2024.
\newblock ISSN 2835-8856.
\newblock URL \url{https://openreview.net/forum?id=ydcrP55u2e}.

\bibitem[Bi et~al.(2025)Bi, Huang, Wang, Yang, Zhang, Huang, Mei, Fang, Li, Wei, Deng, Sun, Zhang, and Liu]{bi2024context}
Baolong Bi, Shaohan Huang, Yiwei Wang, Tianchi Yang, Zihan Zhang, Haizhen Huang, Lingrui Mei, Junfeng Fang, Zehao Li, Furu Wei, Weiwei Deng, Feng Sun, Qi~Zhang, and Shenghua Liu.
\newblock Context-{DPO}: Aligning language models for context-faithfulness.
\newblock In Wanxiang Che, Joyce Nabende, Ekaterina Shutova, and Mohammad~Taher Pilehvar (eds.), \emph{Findings of the Association for Computational Linguistics: ACL 2025}, pp.\  10280--10300, Vienna, Austria, July 2025. Association for Computational Linguistics.
\newblock ISBN 979-8-89176-256-5.
\newblock \doi{10.18653/v1/2025.findings-acl.536}.
\newblock URL \url{https://aclanthology.org/2025.findings-acl.536/}.

\bibitem[Brown et~al.(2020)Brown, Mann, Ryder, Subbiah, Kaplan, Dhariwal, Neelakantan, Shyam, Sastry, Askell, Agarwal, Herbert-Voss, Krueger, Henighan, Child, Ramesh, Ziegler, Wu, Winter, Hesse, Chen, Sigler, Litwin, Gray, Chess, Clark, Berner, McCandlish, Radford, Sutskever, and Amodei]{brown2020language}
Tom Brown, Benjamin Mann, Nick Ryder, Melanie Subbiah, Jared~D Kaplan, Prafulla Dhariwal, Arvind Neelakantan, Pranav Shyam, Girish Sastry, Amanda Askell, Sandhini Agarwal, Ariel Herbert-Voss, Gretchen Krueger, Tom Henighan, Rewon Child, Aditya Ramesh, Daniel Ziegler, Jeffrey Wu, Clemens Winter, Chris Hesse, Mark Chen, Eric Sigler, Mateusz Litwin, Scott Gray, Benjamin Chess, Jack Clark, Christopher Berner, Sam McCandlish, Alec Radford, Ilya Sutskever, and Dario Amodei.
\newblock Language models are few-shot learners.
\newblock In \emph{Advances in Neural Information Processing Systems}, volume~33, pp.\  1877--1901, 2020.

\bibitem[Chan et~al.(2024)Chan, Xu, Yuan, Luo, Xue, Guo, and Fu]{chan2024rqrag}
Chi-Min Chan, Chunpu Xu, Ruibin Yuan, Hongyin Luo, Wei Xue, Yike Guo, and Jie Fu.
\newblock {RQ}-{RAG}: Learning to refine queries for retrieval augmented generation.
\newblock In \emph{First Conference on Language Modeling}, 2024.

\bibitem[Chang et~al.(2025)Chang, Jiang, Rakesh, Pan, Yeh, Wang, Hu, Xu, Zheng, Das, and Zou]{chang2025mainrag}
Chia-Yuan Chang, Zhimeng Jiang, Vineeth Rakesh, Menghai Pan, Chin-Chia~Michael Yeh, Guanchu Wang, Mingzhi Hu, Zhichao Xu, Yan Zheng, Mahashweta Das, and Na~Zou.
\newblock {MAIN}-{RAG}: Multi-agent filtering retrieval-augmented generation.
\newblock In Wanxiang Che, Joyce Nabende, Ekaterina Shutova, and Mohammad~Taher Pilehvar (eds.), \emph{Proceedings of the 63rd Annual Meeting of the Association for Computational Linguistics (Volume 1: Long Papers)}, pp.\  2607--2622, Vienna, Austria, July 2025. Association for Computational Linguistics.
\newblock ISBN 979-8-89176-251-0.
\newblock \doi{10.18653/v1/2025.acl-long.131}.
\newblock URL \url{https://aclanthology.org/2025.acl-long.131/}.

\bibitem[Chann(2023)]{chann2023nondeterminism}
Sherman Chann.
\newblock Non-determinism in gpt-4 is caused by sparse moe, 2023.
\newblock URL \url{https://152334h.github.io/blog/non-determinism-in-gpt-4/}.

\bibitem[Chen et~al.(2025)Chen, Sun, Li, Sun, Zhou, Zhu, Wang, Pan, Zhang, Chen, Yang, Zhou, and Chen]{chen2025research}
Mingyang Chen, Linzhuang Sun, Tianpeng Li, Haoze Sun, Yijie Zhou, Chenzheng Zhu, Haofen Wang, Jeff~Z. Pan, Wen Zhang, Huajun Chen, Fan Yang, Zenan Zhou, and Weipeng Chen.
\newblock Research: Learning to reason with search for llms via reinforcement learning, 2025.
\newblock URL \url{https://arxiv.org/abs/2503.19470}.

\bibitem[Chen et~al.(2024)Chen, Zhong, Ri, Zhao, He, Steinhardt, Yu, and McKeown]{chen2023counterfactualcot}
Yanda Chen, Ruiqi Zhong, Narutatsu Ri, Chen Zhao, He~He, Jacob Steinhardt, Zhou Yu, and Kathleen McKeown.
\newblock Do models explain themselves? counterfactual simulatability of natural language explanations.
\newblock In \emph{Proceedings of the 41st International Conference on Machine Learning}, 2024.

\bibitem[Chu et~al.(2024)Chu, Chen, Chen, Yu, He, Wang, Peng, Liu, Qin, and Liu]{chu2024cotsurvey}
Zheng Chu, Jingchang Chen, Qianglong Chen, Weijiang Yu, Tao He, Haotian Wang, Weihua Peng, Ming Liu, Bing Qin, and Ting Liu.
\newblock Navigate through enigmatic labyrinth a survey of chain of thought reasoning: Advances, frontiers and future.
\newblock In Lun-Wei Ku, Andre Martins, and Vivek Srikumar (eds.), \emph{Proceedings of the 62nd Annual Meeting of the Association for Computational Linguistics (Volume 1: Long Papers)}, pp.\  1173--1203, Bangkok, Thailand, August 2024. Association for Computational Linguistics.
\newblock \doi{10.18653/v1/2024.acl-long.65}.
\newblock URL \url{https://aclanthology.org/2024.acl-long.65/}.

\bibitem[Chua \& Evans(2025)Chua and Evans]{chua2025deepseekfaith}
James Chua and Owain Evans.
\newblock Are deepseek r1 and other reasoning models more faithful?, 2025.
\newblock URL \url{https://arxiv.org/abs/2501.08156}.

\bibitem[Cobbe et~al.(2021)Cobbe, Kosaraju, Bavarian, Chen, Jun, Kaiser, Plappert, Tworek, Hilton, Nakano, Hesse, and Schulman]{cobbe2021trainingverifiers}
Karl Cobbe, Vineet Kosaraju, Mohammad Bavarian, Mark Chen, Heewoo Jun, Lukasz Kaiser, Matthias Plappert, Jerry Tworek, Jacob Hilton, Reiichiro Nakano, Christopher Hesse, and John Schulman.
\newblock Training verifiers to solve math word problems, 2021.
\newblock URL \url{https://arxiv.org/abs/2110.14168}.

\bibitem[Dem{\v{s}}ar(2006)]{demvsar2006statistical}
Janez Dem{\v{s}}ar.
\newblock Statistical comparisons of classifiers over multiple data sets.
\newblock \emph{Journal of Machine learning research}, 7\penalty0 (Jan):\penalty0 1--30, 2006.

\bibitem[Deshpande et~al.(2025)Deshpande, Sirdeshmukh, Mols, Jin, Hernandez-Cardona, Lee, Kritz, Primack, Yue, and Xing]{deshpande-etal-2025-multichallenge}
Kaustubh Deshpande, Ved Sirdeshmukh, Johannes~Baptist Mols, Lifeng Jin, Ed-Yeremai Hernandez-Cardona, Dean Lee, Jeremy Kritz, Willow~E. Primack, Summer Yue, and Chen Xing.
\newblock {M}ulti{C}hallenge: A realistic multi-turn conversation evaluation benchmark challenging to frontier {LLM}s.
\newblock In Wanxiang Che, Joyce Nabende, Ekaterina Shutova, and Mohammad~Taher Pilehvar (eds.), \emph{Findings of the Association for Computational Linguistics: ACL 2025}, pp.\  18632--18702, Vienna, Austria, July 2025. Association for Computational Linguistics.
\newblock ISBN 979-8-89176-256-5.
\newblock \doi{10.18653/v1/2025.findings-acl.958}.
\newblock URL \url{https://aclanthology.org/2025.findings-acl.958/}.

\bibitem[Dror et~al.(2018)Dror, Baumer, Shlomov, and Reichart]{dror-etal-2018-hitchhikers}
Rotem Dror, Gili Baumer, Segev Shlomov, and Roi Reichart.
\newblock The hitchhiker{'}s guide to testing statistical significance in natural language processing.
\newblock In Iryna Gurevych and Yusuke Miyao (eds.), \emph{Proceedings of the 56th Annual Meeting of the Association for Computational Linguistics (Volume 1: Long Papers)}, pp.\  1383--1392, Melbourne, Australia, July 2018. Association for Computational Linguistics.
\newblock \doi{10.18653/v1/P18-1128}.
\newblock URL \url{https://aclanthology.org/P18-1128/}.

\bibitem[Dziri et~al.(2022{\natexlab{a}})Dziri, Kamalloo, Milton, Zaiane, Yu, Ponti, and Reddy]{dziri2022faithdial}
Nouha Dziri, Ehsan Kamalloo, Sivan Milton, Osmar Zaiane, Mo~Yu, Edoardo~M. Ponti, and Siva Reddy.
\newblock {F}aith{D}ial: A faithful benchmark for information-seeking dialogue.
\newblock \emph{Transactions of the Association for Computational Linguistics}, 10:\penalty0 1473--1490, 2022{\natexlab{a}}.
\newblock \doi{10.1162/tacl_a_00529}.
\newblock URL \url{https://aclanthology.org/2022.tacl-1.84/}.

\bibitem[Dziri et~al.(2022{\natexlab{b}})Dziri, Milton, Yu, Zaiane, and Reddy]{dziri2022origincontextfaith}
Nouha Dziri, Sivan Milton, Mo~Yu, Osmar Zaiane, and Siva Reddy.
\newblock On the origin of hallucinations in conversational models: Is it the datasets or the models?
\newblock In Marine Carpuat, Marie-Catherine de~Marneffe, and Ivan~Vladimir Meza~Ruiz (eds.), \emph{Proceedings of the 2022 Conference of the North American Chapter of the Association for Computational Linguistics: Human Language Technologies}, pp.\  5271--5285, Seattle, United States, July 2022{\natexlab{b}}. Association for Computational Linguistics.
\newblock \doi{10.18653/v1/2022.naacl-main.387}.
\newblock URL \url{https://aclanthology.org/2022.naacl-main.387/}.

\bibitem[Fan et~al.(2024)Fan, Ding, Ning, Wang, Li, Yin, Chua, and Li]{fan2024survey}
Wenqi Fan, Yujuan Ding, Liangbo Ning, Shijie Wang, Hengyun Li, Dawei Yin, Tat-Seng Chua, and Qing Li.
\newblock A survey on rag meeting llms: Towards retrieval-augmented large language models.
\newblock In \emph{Proceedings of the 30th ACM SIGKDD conference on knowledge discovery and data mining}, pp.\  6491--6501, 2024.

\bibitem[Gao et~al.(2023)Gao, Xiong, Gao, Jia, Pan, Bi, Dai, Sun, Wang, and Wang]{gao2023ragsurvey}
Yunfan Gao, Yun Xiong, Xinyu Gao, Kangxiang Jia, Jinliu Pan, Yuxi Bi, Yi~Dai, Jiawei Sun, Meng Wang, and Haofen Wang.
\newblock Retrieval-augmented generation for large language models: A survey, 2023.
\newblock URL \url{https://arxiv.org/abs/2312.10997}.

\bibitem[Georgiev et~al.(2024)Georgiev, Lei, Burnell, Bai, Gulati, Tanzer, Vincent, Pan, Wang, et~al.]{google2024gemini}
Petko Georgiev, Ving~Ian Lei, Ryan Burnell, Libin Bai, Anmol Gulati, Garrett Tanzer, Damien Vincent, Zhufeng Pan, Shibo Wang, et~al.
\newblock Gemini 1.5: Unlocking multimodal understanding across millions of tokens of context, 2024.
\newblock URL \url{https://arxiv.org/abs/2403.05530}.

\bibitem[Geva et~al.(2021)Geva, Khashabi, Segal, Khot, Roth, and Berant]{Geva2021qareasonbenchmark}
Mor Geva, Daniel Khashabi, Elad Segal, Tushar Khot, Dan Roth, and Jonathan Berant.
\newblock Did aristotle use a laptop? a question answering benchmark with implicit reasoning strategies.
\newblock \emph{Transactions of the Association for Computational Linguistics}, 9:\penalty0 346--361, 2021.
\newblock \doi{10.1162/tacl_a_00370}.
\newblock URL \url{https://aclanthology.org/2021.tacl-1.21/}.

\bibitem[Gheshlaghi~Azar et~al.(2024)Gheshlaghi~Azar, Daniel~Guo, Piot, Munos, Rowland, Valko, and Calandriello]{pmlr-v238-gheshlaghi-azar24a}
Mohammad Gheshlaghi~Azar, Zhaohan Daniel~Guo, Bilal Piot, Remi Munos, Mark Rowland, Michal Valko, and Daniele Calandriello.
\newblock A general theoretical paradigm to understand learning from human preferences.
\newblock In \emph{Proceedings of The 27th International Conference on Artificial Intelligence and Statistics}, pp.\  4447--4455. PMLR, 2024.

\bibitem[Grattafiori et~al.(2024)Grattafiori, Dubey, Jauhri, Pandey, Kadian, Al-Dahle, Letman, Mathur, Schelten, et~al.]{grattafiori2024llama3}
Aaron Grattafiori, Abhimanyu Dubey, Abhinav Jauhri, Abhinav Pandey, Abhishek Kadian, Ahmad Al-Dahle, Aiesha Letman, Akhil Mathur, Alan Schelten, et~al.
\newblock The llama 3 herd of models, 2024.
\newblock URL \url{https://arxiv.org/abs/2407.21783}.

\bibitem[Gunjal et~al.(2026)Gunjal, Wang, Lau, Nath, He, Liu, and Hendryx]{gunjal2026rubricsasrewards}
Anisha Gunjal, Anthony Wang, Elaine Lau, Vaskar Nath, Yunzhong He, Bing Liu, and Sean~M. Hendryx.
\newblock Rubrics as rewards: Reinforcement learning beyond verifiable domains.
\newblock In \emph{The Fourteenth International Conference on Learning Representations}, 2026.
\newblock URL \url{https://openreview.net/forum?id=c1bTcrDmt4}.

\bibitem[Guo et~al.(2025)Guo, Yang, Zhang, Song, Wang, Zhu, Xu, Zhang, Ma, Bi, et~al.]{guo2025deepseek}
Daya Guo, Dejian Yang, Haowei Zhang, Junxiao Song, Peiyi Wang, Qihao Zhu, Runxin Xu, Ruoyu Zhang, Shirong Ma, Xiao Bi, et~al.
\newblock Deepseek-r1 incentivizes reasoning in llms through reinforcement learning.
\newblock \emph{Nature}, 645:\penalty0 633–638, 2025.
\newblock ISSN 1476-4687.
\newblock URL \url{http://dx.doi.org/10.1038/s41586-025-09422-z}.

\bibitem[He \& Lab(2025)He and Lab]{he2025nondeterminism}
Horace He and Thinking~Machines Lab.
\newblock Defeating nondeterminism in llm inference.
\newblock \emph{Thinking Machines Lab: Connectionism}, 2025.
\newblock \doi{10.64434/tml.20250910}.
\newblock URL \url{https://thinkingmachines.ai/blog/defeating-nondeterminism-in-llm-inference/}.

\bibitem[Henderson et~al.(2018)Henderson, Islam, Bachman, Pineau, Precup, and Meger]{henderson-etal-2018-deepreinforcement}
Peter Henderson, Riashat Islam, Philip Bachman, Joelle Pineau, Doina Precup, and David Meger.
\newblock Deep reinforcement learning that matters.
\newblock In \emph{Proceedings of the Thirty-Second AAAI Conference on Artificial Intelligence and Thirtieth Innovative Applications of Artificial Intelligence Conference and Eighth AAAI Symposium on Educational Advances in Artificial Intelligence}, AAAI'18/IAAI'18/EAAI'18. AAAI Press, 2018.
\newblock ISBN 978-1-57735-800-8.

\bibitem[Ho et~al.(2020)Ho, Duong~Nguyen, Sugawara, and Aizawa]{ho-etal-2020-constructing}
Xanh Ho, Anh-Khoa Duong~Nguyen, Saku Sugawara, and Akiko Aizawa.
\newblock Constructing a multi-hop {QA} dataset for comprehensive evaluation of reasoning steps.
\newblock In \emph{Proceedings of the 28th International Conference on Computational Linguistics}, pp.\  6609--6625, 2020.

\bibitem[Honovich et~al.(2022)Honovich, Aharoni, Herzig, Taitelbaum, Kukliansy, Cohen, Scialom, Szpektor, Hassidim, and Matias]{honovich-etal-2022-true-evaluating}
Or~Honovich, Roee Aharoni, Jonathan Herzig, Hagai Taitelbaum, Doron Kukliansy, Vered Cohen, Thomas Scialom, Idan Szpektor, Avinatan Hassidim, and Yossi Matias.
\newblock {TRUE}: Re-evaluating factual consistency evaluation.
\newblock In Marine Carpuat, Marie-Catherine de~Marneffe, and Ivan~Vladimir Meza~Ruiz (eds.), \emph{Proceedings of the 2022 Conference of the North American Chapter of the Association for Computational Linguistics: Human Language Technologies}, pp.\  3905--3920, Seattle, United States, July 2022. Association for Computational Linguistics.
\newblock \doi{10.18653/v1/2022.naacl-main.287}.
\newblock URL \url{https://aclanthology.org/2022.naacl-main.287/}.

\bibitem[Hu et~al.(2022)Hu, yelong shen, Wallis, Allen-Zhu, Li, Wang, Wang, and Chen]{hu2022lora}
Edward~J Hu, yelong shen, Phillip Wallis, Zeyuan Allen-Zhu, Yuanzhi Li, Shean Wang, Lu~Wang, and Weizhu Chen.
\newblock Lo{RA}: Low-rank adaptation of large language models.
\newblock In \emph{International Conference on Learning Representations}, 2022.
\newblock URL \url{https://openreview.net/forum?id=nZeVKeeFYf9}.

\bibitem[Huang et~al.(2025)Huang, Yu, Ma, Zhong, Feng, Wang, Chen, Peng, Feng, Qin, and Liu]{Huang2025SurveyOnHallucination}
Lei Huang, Weijiang Yu, Weitao Ma, Weihong Zhong, Zhangyin Feng, Haotian Wang, Qianglong Chen, Weihua Peng, Xiaocheng Feng, Bing Qin, and Ting Liu.
\newblock A survey on hallucination in large language models: Principles, taxonomy, challenges, and open questions.
\newblock \emph{ACM Transactions on Information Systems}, 2025.

\bibitem[Jaech et~al.(2024)Jaech, Kalai, Lerer, Richardson, El-Kishky, Low, Helyar, Madry, Beutel, Carney, et~al.]{jaech2024openai}
Aaron Jaech, Adam Kalai, Adam Lerer, Adam Richardson, Ahmed El-Kishky, Aiden Low, Alec Helyar, Aleksander Madry, Alex Beutel, Alex Carney, et~al.
\newblock Openai o1 system card, 2024.
\newblock URL \url{https://arxiv.org/abs/2412.16720}.

\bibitem[Jiang et~al.(2024)Jiang, Xie, Hao, Wang, Mallick, Su, Taylor, and Roth]{jiang-etal-2024-peek}
Bowen Jiang, Yangxinyu Xie, Zhuoqun Hao, Xiaomeng Wang, Tanwi Mallick, Weijie~J Su, Camillo~Jose Taylor, and Dan Roth.
\newblock A peek into token bias: Large language models are not yet genuine reasoners.
\newblock In Yaser Al-Onaizan, Mohit Bansal, and Yun-Nung Chen (eds.), \emph{Proceedings of the 2024 Conference on Empirical Methods in Natural Language Processing}, pp.\  4722--4756, Miami, Florida, USA, November 2024. Association for Computational Linguistics.
\newblock \doi{10.18653/v1/2024.emnlp-main.272}.
\newblock URL \url{https://aclanthology.org/2024.emnlp-main.272/}.

\bibitem[Jiang et~al.(2025)Jiang, Chen, Li, Ren, Wang, Zhao, Song, and Zhang]{jiang2025ragstar}
Jinhao Jiang, Jiayi Chen, Junyi Li, Ruiyang Ren, Shijie Wang, Wayne~Xin Zhao, Yang Song, and Tao Zhang.
\newblock {RAG}-star: Enhancing deliberative reasoning with retrieval augmented verification and refinement.
\newblock In Luis Chiruzzo, Alan Ritter, and Lu~Wang (eds.), \emph{Proceedings of the 2025 Conference of the Nations of the Americas Chapter of the Association for Computational Linguistics: Human Language Technologies (Volume 1: Long Papers)}, pp.\  7064--7074, Albuquerque, New Mexico, April 2025. Association for Computational Linguistics.
\newblock ISBN 979-8-89176-189-6.
\newblock \doi{10.18653/v1/2025.naacl-long.361}.
\newblock URL \url{https://aclanthology.org/2025.naacl-long.361/}.

\bibitem[Jiang et~al.(2023)Jiang, Xu, Gao, Sun, Liu, Dwivedi-Yu, Yang, Callan, and Neubig]{jiang2023activerag}
Zhengbao Jiang, Frank Xu, Luyu Gao, Zhiqing Sun, Qian Liu, Jane Dwivedi-Yu, Yiming Yang, Jamie Callan, and Graham Neubig.
\newblock Active retrieval augmented generation.
\newblock In Houda Bouamor, Juan Pino, and Kalika Bali (eds.), \emph{Proceedings of the 2023 Conference on Empirical Methods in Natural Language Processing}, pp.\  7969--7992, Singapore, December 2023. Association for Computational Linguistics.
\newblock \doi{10.18653/v1/2023.emnlp-main.495}.
\newblock URL \url{https://aclanthology.org/2023.emnlp-main.495/}.

\bibitem[Jin et~al.(2025{\natexlab{a}})Jin, Yoon, Kargupta, Arik, and Han]{jin2025searchr1extension}
Bowen Jin, Jinsung Yoon, Priyanka Kargupta, Sercan~O. Arik, and Jiawei Han.
\newblock An empirical study on reinforcement learning for reasoning-search interleaved llm agents, 2025{\natexlab{a}}.
\newblock URL \url{https://arxiv.org/abs/2505.15117}.

\bibitem[Jin et~al.(2025{\natexlab{b}})Jin, Zeng, Yue, Yoon, Arik, Wang, Zamani, and Han]{jin2025searchr1}
Bowen Jin, Hansi Zeng, Zhenrui Yue, Jinsung Yoon, Sercan~O Arik, Dong Wang, Hamed Zamani, and Jiawei Han.
\newblock Search-r1: Training {LLM}s to reason and leverage search engines with reinforcement learning.
\newblock In \emph{Second Conference on Language Modeling}, 2025{\natexlab{b}}.
\newblock URL \url{https://openreview.net/forum?id=Rwhi91ideu}.

\bibitem[Jin et~al.(2025{\natexlab{c}})Jin, Zhu, Dou, Dong, Yang, Zhang, Zhao, Yang, and Wen]{jin2025FlashRAG}
Jiajie Jin, Yutao Zhu, Zhicheng Dou, Guanting Dong, Xinyu Yang, Chenghao Zhang, Tong Zhao, Zhao Yang, and Ji-Rong Wen.
\newblock Flashrag: A modular toolkit for efficient retrieval-augmented generation research.
\newblock In \emph{Companion Proceedings of the ACM on Web Conference 2025}, pp.\  737–740, 2025{\natexlab{c}}.

\bibitem[Joshi et~al.(2017)Joshi, Choi, Weld, and Zettlemoyer]{joshi2017triviaqa}
Mandar Joshi, Eunsol Choi, Daniel~S. Weld, and Luke Zettlemoyer.
\newblock Triviaqa: A large scale distantly supervised challenge dataset for reading comprehension, 2017.
\newblock URL \url{https://arxiv.org/abs/1705.03551}.

\bibitem[Kaelbling et~al.(1996)Kaelbling, Littman, and Moore]{kaelbling1996reinforcement}
Leslie~Pack Kaelbling, Michael~L Littman, and Andrew~W Moore.
\newblock Reinforcement learning: A survey.
\newblock \emph{Journal of artificial intelligence research}, 4:\penalty0 237--285, 1996.

\bibitem[Kaufmann et~al.(2025)Kaufmann, Weng, Bengs, and H{\"u}llermeier]{kaufmann2025a}
Timo Kaufmann, Paul Weng, Viktor Bengs, and Eyke H{\"u}llermeier.
\newblock A survey of reinforcement learning from human feedback.
\newblock \emph{Transactions on Machine Learning Research}, 2025.
\newblock ISSN 2835-8856.

\bibitem[Khan et~al.(2024)Khan, Hughes, Valentine, Ruis, Sachan, Radhakrishnan, Grefenstette, Bowman, Rockt\"{a}schel, and Perez]{khan2024debatingwithmorepersuasive}
Akbir Khan, John Hughes, Dan Valentine, Laura Ruis, Kshitij Sachan, Ansh Radhakrishnan, Edward Grefenstette, Samuel~R. Bowman, Tim Rockt\"{a}schel, and Ethan Perez.
\newblock Debating with more persuasive {LLM}s leads to more truthful answers.
\newblock In Ruslan Salakhutdinov, Zico Kolter, Katherine Heller, Adrian Weller, Nuria Oliver, Jonathan Scarlett, and Felix Berkenkamp (eds.), \emph{Proceedings of the 41st International Conference on Machine Learning}, volume 235 of \emph{Proceedings of Machine Learning Research}, pp.\  23662--23733. PMLR, 21--27 Jul 2024.
\newblock URL \url{https://proceedings.mlr.press/v235/khan24a.html}.

\bibitem[Kojima et~al.(2022)Kojima, Gu, Reid, Matsuo, and Iwasawa]{kojima2022largelanguagemodelsarezeroshotreasoners}
Takeshi Kojima, Shixiang~(Shane) Gu, Machel Reid, Yutaka Matsuo, and Yusuke Iwasawa.
\newblock Large language models are zero-shot reasoners.
\newblock In \emph{Advances in Neural Information Processing Systems}, volume~35, pp.\  22199--22213, 2022.

\bibitem[Kool et~al.(2019)Kool, van Hoof, and Welling]{kool2019buyfoursamples}
Wouter Kool, Herke van Hoof, and Max Welling.
\newblock Buy 4 {REINFORCE} samples, get a baseline for free!, 2019.
\newblock URL \url{https://openreview.net/forum?id=r1lgTGL5DE}.

\bibitem[Kwiatkowski et~al.(2019)Kwiatkowski, Palomaki, Redfield, Collins, Parikh, Alberti, Epstein, Polosukhin, Devlin, Lee, Toutanova, Jones, Kelcey, Chang, Dai, Uszkoreit, Le, and Petrov]{kwiatkowski2019nqdata}
Tom Kwiatkowski, Jennimaria Palomaki, Olivia Redfield, Michael Collins, Ankur Parikh, Chris Alberti, Danielle Epstein, Illia Polosukhin, Jacob Devlin, Kenton Lee, Kristina Toutanova, Llion Jones, Matthew Kelcey, Ming-Wei Chang, Andrew~M. Dai, Jakob Uszkoreit, Quoc Le, and Slav Petrov.
\newblock Natural questions: A benchmark for question answering research.
\newblock \emph{Transactions of the Association for Computational Linguistics}, 7:\penalty0 452--466, 2019.
\newblock \doi{10.1162/tacl_a_00276}.
\newblock URL \url{https://aclanthology.org/Q19-1026/}.

\bibitem[Lanchantin et~al.(2025)Lanchantin, Chen, Lan, Li, Saha, Wang, Xu, Yu, Yuan, Weston, Sukhbaatar, and Kulikov]{lanchantin2025bridging}
Jack Lanchantin, Angelica Chen, Janice Lan, Xian Li, Swarnadeep Saha, Tianlu Wang, Jing Xu, Ping Yu, Weizhe Yuan, Jason~E Weston, Sainbayar Sukhbaatar, and Ilia Kulikov.
\newblock Bridging offline and online reinforcement learning for llms, 2025.
\newblock URL \url{https://arxiv.org/abs/2506.21495}.

\bibitem[Lanham et~al.(2023)Lanham, Chen, Radhakrishnan, Steiner, Denison, Hernandez, Li, Durmus, Hubinger, Kernion, Lukošiūtė, Nguyen, Cheng, Joseph, Schiefer, Rausch, Larson, McCandlish, Kundu, Kadavath, Yang, Henighan, Maxwell, Telleen-Lawton, Hume, Hatfield-Dodds, Kaplan, Brauner, Bowman, and Perez]{Lanham2023MeasuringCotFaith}
Tamera Lanham, Anna Chen, Ansh Radhakrishnan, Benoit Steiner, Carson Denison, Danny Hernandez, Dustin Li, Esin Durmus, Evan Hubinger, Jackson Kernion, Kamilė Lukošiūtė, Karina Nguyen, Newton Cheng, Nicholas Joseph, Nicholas Schiefer, Oliver Rausch, Robin Larson, Sam McCandlish, Sandipan Kundu, Saurav Kadavath, Shannon Yang, Thomas Henighan, Timothy Maxwell, Timothy Telleen-Lawton, Tristan Hume, Zac Hatfield-Dodds, Jared Kaplan, Jan Brauner, Samuel~R. Bowman, and Ethan Perez.
\newblock Measuring faithfulness in chain-of-thought reasoning, 2023.
\newblock URL \url{https://arxiv.org/abs/2307.13702}.

\bibitem[Lee et~al.(2025)Lee, Zeng, Jeong, yong Sohn, and Lee]{lee2026correctlyreportllmasajudgeevaluations}
Chungpa Lee, Thomas Zeng, Jongwon Jeong, Jy~yong Sohn, and Kangwook Lee.
\newblock How to correctly report llm-as-a-judge evaluations, 2025.
\newblock URL \url{https://arxiv.org/abs/2511.21140}.

\bibitem[Lewis et~al.(2020)Lewis, Perez, Piktus, Petroni, Karpukhin, Goyal, K\"{u}ttler, Lewis, Yih, Rockt\"{a}schel, Riedel, and Kiela]{Patrick2020rag}
Patrick Lewis, Ethan Perez, Aleksandra Piktus, Fabio Petroni, Vladimir Karpukhin, Naman Goyal, Heinrich K\"{u}ttler, Mike Lewis, Wen-tau Yih, Tim Rockt\"{a}schel, Sebastian Riedel, and Douwe Kiela.
\newblock Retrieval-augmented generation for knowledge-intensive nlp tasks.
\newblock In \emph{Advances in Neural Information Processing Systems}, pp.\  9459--9474, 2020.

\bibitem[Li et~al.(2024)Li, Dong, Chen, Su, Zhou, Ai, Ye, and Liu]{li2024llmsasjudgescomprehensivesurvey}
Haitao Li, Qian Dong, Junjie Chen, Huixue Su, Yujia Zhou, Qingyao Ai, Ziyi Ye, and Yiqun Liu.
\newblock Llms-as-judges: A comprehensive survey on llm-based evaluation methods, 2024.
\newblock URL \url{https://arxiv.org/abs/2412.05579}.

\bibitem[Li et~al.(2023)Li, Cheng, Zhao, Nie, and Wen]{li2023halueval}
Junyi Li, Xiaoxue Cheng, Xin Zhao, Jian-Yun Nie, and Ji-Rong Wen.
\newblock {H}alu{E}val: A large-scale hallucination evaluation benchmark for large language models.
\newblock In Houda Bouamor, Juan Pino, and Kalika Bali (eds.), \emph{Proceedings of the 2023 Conference on Empirical Methods in Natural Language Processing}, pp.\  6449--6464, Singapore, December 2023. Association for Computational Linguistics.
\newblock \doi{10.18653/v1/2023.emnlp-main.397}.
\newblock URL \url{https://aclanthology.org/2023.emnlp-main.397/}.

\bibitem[Li et~al.(2025)Li, Dong, Jin, Zhang, Zhou, Zhu, Zhang, and Dou]{li2025searcho1}
Xiaoxi Li, Guanting Dong, Jiajie Jin, Yuyao Zhang, Yujia Zhou, Yutao Zhu, Peitian Zhang, and Zhicheng Dou.
\newblock Search-o1: Agentic search-enhanced large reasoning models.
\newblock In Christos Christodoulopoulos, Tanmoy Chakraborty, Carolyn Rose, and Violet Peng (eds.), \emph{Proceedings of the 2025 Conference on Empirical Methods in Natural Language Processing}, pp.\  5420--5438, Suzhou, China, November 2025. Association for Computational Linguistics.
\newblock ISBN 979-8-89176-332-6.
\newblock \doi{10.18653/v1/2025.emnlp-main.276}.
\newblock URL \url{https://aclanthology.org/2025.emnlp-main.276/}.

\bibitem[Liang et~al.(2025)Liang, Sugang, Lin, Wu, Zhao, and Li]{liang2025reasoningrag}
Jintao Liang, Sugang, Huifeng Lin, You Wu, Rui Zhao, and Ziyue Li.
\newblock Reasoning {RAG} via system 1 or system 2: A survey on reasoning agentic retrieval-augmented generation for industry challenges.
\newblock In Kentaro Inui, Sakriani Sakti, Haofen Wang, Derek~F. Wong, Pushpak Bhattacharyya, Biplab Banerjee, Asif Ekbal, Tanmoy Chakraborty, and Dhirendra~Pratap Singh (eds.), \emph{Proceedings of the 14th International Joint Conference on Natural Language Processing and the 4th Conference of the Asia-Pacific Chapter of the Association for Computational Linguistics}, pp.\  1954--1966, Mumbai, India, December 2025. The Asian Federation of Natural Language Processing and The Association for Computational Linguistics.
\newblock ISBN 979-8-89176-303-6.
\newblock \doi{10.18653/v1/2025.findings-ijcnlp.122}.
\newblock URL \url{https://aclanthology.org/2025.findings-ijcnlp.122/}.

\bibitem[Lightman et~al.(2024)Lightman, Kosaraju, Burda, Edwards, Baker, Lee, Leike, Schulman, Sutskever, and Cobbe]{lightman2023letsverify}
Hunter Lightman, Vineet Kosaraju, Yuri Burda, Harrison Edwards, Bowen Baker, Teddy Lee, Jan Leike, John Schulman, Ilya Sutskever, and Karl Cobbe.
\newblock Let's verify step by step.
\newblock In \emph{The Twelfth International Conference on Learning Representations}, 2024.
\newblock URL \url{https://openreview.net/forum?id=v8L0pN6EOi}.

\bibitem[Lin et~al.(2025)Lin, Wu, Xu, Liu, Tang, He, Aggarwal, Liu, Zhang, and Wang]{lin2025comprehensivesurveyreinforcementlearningbased}
Minhua Lin, Zongyu Wu, Zhichao Xu, Hui Liu, Xianfeng Tang, Qi~He, Charu Aggarwal, Hui Liu, Xiang Zhang, and Suhang Wang.
\newblock A comprehensive survey on reinforcement learning-based agentic search: Foundations, roles, optimizations, evaluations, and applications, 2025.
\newblock URL \url{https://arxiv.org/abs/2510.16724}.

\bibitem[Ma et~al.(2023)Ma, Gong, He, Zhao, and Duan]{ma2023queryRewriting}
Xinbei Ma, Yeyun Gong, Pengcheng He, Hai Zhao, and Nan Duan.
\newblock Query rewriting in retrieval-augmented large language models.
\newblock In Houda Bouamor, Juan Pino, and Kalika Bali (eds.), \emph{Proceedings of the 2023 Conference on Empirical Methods in Natural Language Processing}, pp.\  5303--5315, Singapore, December 2023. Association for Computational Linguistics.
\newblock \doi{10.18653/v1/2023.emnlp-main.322}.
\newblock URL \url{https://aclanthology.org/2023.emnlp-main.322/}.

\bibitem[Malaviya et~al.(2025)Malaviya, Chang, Roth, Iyyer, Yatskar, and Lo]{malaviya2025contextualized}
Chaitanya Malaviya, Joseph~Chee Chang, Dan Roth, Mohit Iyyer, Mark Yatskar, and Kyle Lo.
\newblock Contextualized evaluations: Judging language model responses to underspecified queries.
\newblock \emph{Transactions of the Association for Computational Linguistics}, 13:\penalty0 878--900, 2025.
\newblock \doi{10.1162/tacl.a.24}.
\newblock URL \url{https://aclanthology.org/2025.tacl-1.41/}.

\bibitem[Mallen et~al.(2023)Mallen, Asai, Zhong, Das, Khashabi, and Hajishirzi]{mallen-etal-2023-whennottotrust}
Alex Mallen, Akari Asai, Victor Zhong, Rajarshi Das, Daniel Khashabi, and Hannaneh Hajishirzi.
\newblock When not to trust language models: Investigating effectiveness of parametric and non-parametric memories.
\newblock In Anna Rogers, Jordan Boyd-Graber, and Naoaki Okazaki (eds.), \emph{Proceedings of the 61st Annual Meeting of the Association for Computational Linguistics (Volume 1: Long Papers)}, pp.\  9802--9822, Toronto, Canada, July 2023. Association for Computational Linguistics.
\newblock \doi{10.18653/v1/2023.acl-long.546}.
\newblock URL \url{https://aclanthology.org/2023.acl-long.546/}.

\bibitem[Marchionini(2006)]{marchionini2006exploratory}
Gary Marchionini.
\newblock Exploratory search: from finding to understanding.
\newblock \emph{Communications of the ACM}, 49\penalty0 (4):\penalty0 41--46, 2006.

\bibitem[Matton et~al.(2025)Matton, Ness, Guttag, and Kiciman]{matton2025faithmeasure}
Katie Matton, Robert Ness, John Guttag, and Emre Kiciman.
\newblock Walk the talk? measuring the faithfulness of large language model explanations.
\newblock In \emph{The Thirteenth International Conference on Learning Representations}, 2025.

\bibitem[Maynez et~al.(2020)Maynez, Narayan, Bohnet, and McDonald]{maynez-etal-2020-faithfulness}
Joshua Maynez, Shashi Narayan, Bernd Bohnet, and Ryan McDonald.
\newblock On faithfulness and factuality in abstractive summarization.
\newblock In Dan Jurafsky, Joyce Chai, Natalie Schluter, and Joel Tetreault (eds.), \emph{Proceedings of the 58th Annual Meeting of the Association for Computational Linguistics}, pp.\  1906--1919, Online, July 2020. Association for Computational Linguistics.
\newblock \doi{10.18653/v1/2020.acl-main.173}.
\newblock URL \url{https://aclanthology.org/2020.acl-main.173/}.

\bibitem[Meng et~al.(2024)Meng, Xia, and Chen]{meng2024simpo}
Yu~Meng, Mengzhou Xia, and Danqi Chen.
\newblock Sim{PO}: Simple preference optimization with a reference-free reward.
\newblock In \emph{The Thirty-eighth Annual Conference on Neural Information Processing Systems}, 2024.

\bibitem[Ming et~al.(2025)Ming, Purushwalkam, Pandit, Ke, Nguyen, Xiong, and Joty]{ming2025faitheval}
Yifei Ming, Senthil Purushwalkam, Shrey Pandit, Zixuan Ke, Xuan-Phi Nguyen, Caiming Xiong, and Shafiq Joty.
\newblock Faitheval: Can your language model stay faithful to context, even if ''the moon is made of marshmallows''.
\newblock In \emph{The Thirteenth International Conference on Learning Representations}, 2025.

\bibitem[Mousavi et~al.(2024)Mousavi, Alghisi, and Riccardi]{mousavi2024dyknow}
Seyed~Mahed Mousavi, Simone Alghisi, and Giuseppe Riccardi.
\newblock {D}y{K}now: Dynamically verifying time-sensitive factual knowledge in {LLM}s.
\newblock In Yaser Al-Onaizan, Mohit Bansal, and Yun-Nung Chen (eds.), \emph{Findings of the Association for Computational Linguistics: EMNLP 2024}, pp.\  8014--8029, Miami, Florida, USA, November 2024. Association for Computational Linguistics.
\newblock \doi{10.18653/v1/2024.findings-emnlp.471}.
\newblock URL \url{https://aclanthology.org/2024.findings-emnlp.471/}.

\bibitem[Ouyang et~al.(2022)Ouyang, Wu, Jiang, Almeida, Wainwright, Mishkin, Zhang, Agarwal, Slama, Ray, Schulman, Hilton, Kelton, Miller, Simens, Askell, Welinder, Christiano, Leike, and Lowe]{ouyang2022training}
Long Ouyang, Jeffrey Wu, Xu~Jiang, Diogo Almeida, Carroll Wainwright, Pamela Mishkin, Chong Zhang, Sandhini Agarwal, Katarina Slama, Alex Ray, John Schulman, Jacob Hilton, Fraser Kelton, Luke Miller, Maddie Simens, Amanda Askell, Peter Welinder, Paul~F Christiano, Jan Leike, and Ryan Lowe.
\newblock Training language models to follow instructions with human feedback.
\newblock In \emph{Advances in Neural Information Processing Systems}, volume~35, pp.\  27730--27744, 2022.

\bibitem[Parcalabescu \& Frank(2024)Parcalabescu and Frank]{parcalabescu2024measuring}
Letitia Parcalabescu and Anette Frank.
\newblock On measuring faithfulness or self-consistency of natural language explanations.
\newblock In Lun-Wei Ku, Andre Martins, and Vivek Srikumar (eds.), \emph{Proceedings of the 62nd Annual Meeting of the Association for Computational Linguistics (Volume 1: Long Papers)}, pp.\  6048--6089, Bangkok, Thailand, August 2024. Association for Computational Linguistics.
\newblock \doi{10.18653/v1/2024.acl-long.329}.
\newblock URL \url{https://aclanthology.org/2024.acl-long.329/}.

\bibitem[Patil et~al.(2024)Patil, Zhang, Wang, and Gonzalez]{patil2024gorilla}
Shishir~G Patil, Tianjun Zhang, Xin Wang, and Joseph~E. Gonzalez.
\newblock Gorilla: Large language model connected with massive {API}s.
\newblock In \emph{The Thirty-eighth Annual Conference on Neural Information Processing Systems}, 2024.
\newblock URL \url{https://openreview.net/forum?id=tBRNC6YemY}.

\bibitem[Paul et~al.(2024)Paul, West, Bosselut, and Faltings]{paul-etal-2024-making}
Debjit Paul, Robert West, Antoine Bosselut, and Boi Faltings.
\newblock Making reasoning matter: Measuring and improving faithfulness of chain-of-thought reasoning.
\newblock In Yaser Al-Onaizan, Mohit Bansal, and Yun-Nung Chen (eds.), \emph{Findings of the Association for Computational Linguistics: EMNLP 2024}, pp.\  15012--15032, Miami, Florida, USA, November 2024. Association for Computational Linguistics.
\newblock \doi{10.18653/v1/2024.findings-emnlp.882}.
\newblock URL \url{https://aclanthology.org/2024.findings-emnlp.882/}.

\bibitem[Press et~al.(2023)Press, Zhang, Min, Schmidt, Smith, and Lewis]{press2023bamboogle}
Ofir Press, Muru Zhang, Sewon Min, Ludwig Schmidt, Noah Smith, and Mike Lewis.
\newblock Measuring and narrowing the compositionality gap in language models.
\newblock In Houda Bouamor, Juan Pino, and Kalika Bali (eds.), \emph{Findings of the Association for Computational Linguistics: EMNLP 2023}, pp.\  5687--5711, Singapore, December 2023. Association for Computational Linguistics.
\newblock \doi{10.18653/v1/2023.findings-emnlp.378}.
\newblock URL \url{https://aclanthology.org/2023.findings-emnlp.378/}.

\bibitem[Rafailov et~al.(2023)Rafailov, Sharma, Mitchell, Manning, Ermon, and Finn]{rafailov2023direct}
Rafael Rafailov, Archit Sharma, Eric Mitchell, Christopher~D Manning, Stefano Ermon, and Chelsea Finn.
\newblock Direct preference optimization: Your language model is secretly a reward model.
\newblock In \emph{Advances in Neural Information Processing Systems}, volume~36, pp.\  53728--53741, 2023.

\bibitem[Ribeiro et~al.(2020)Ribeiro, Wu, Guestrin, and Singh]{ribeiro-etal-2020-beyond}
Marco~Tulio Ribeiro, Tongshuang Wu, Carlos Guestrin, and Sameer Singh.
\newblock Beyond accuracy: Behavioral testing of {NLP} models with {C}heck{L}ist.
\newblock In Dan Jurafsky, Joyce Chai, Natalie Schluter, and Joel Tetreault (eds.), \emph{Proceedings of the 58th Annual Meeting of the Association for Computational Linguistics}, pp.\  4902--4912, Online, July 2020. Association for Computational Linguistics.
\newblock \doi{10.18653/v1/2020.acl-main.442}.
\newblock URL \url{https://aclanthology.org/2020.acl-main.442/}.

\bibitem[Schick et~al.(2023)Schick, Dwivedi-Yu, Dessi, Raileanu, Lomeli, Hambro, Zettlemoyer, Cancedda, and Scialom]{schick2023toolformer}
Timo Schick, Jane Dwivedi-Yu, Roberto Dessi, Roberta Raileanu, Maria Lomeli, Eric Hambro, Luke Zettlemoyer, Nicola Cancedda, and Thomas Scialom.
\newblock Toolformer: Language models can teach themselves to use tools.
\newblock In \emph{Thirty-seventh Conference on Neural Information Processing Systems}, 2023.

\bibitem[Schulman et~al.(2015)Schulman, Moritz, Levine, Jordan, and Abbeel]{schulman2015high}
John Schulman, Philipp Moritz, Sergey Levine, Michael Jordan, and Pieter Abbeel.
\newblock High-dimensional continuous control using generalized advantage estimation, 2015.
\newblock URL \url{https://arxiv.org/abs/1506.02438}.

\bibitem[Schulman et~al.(2017)Schulman, Wolski, Dhariwal, Radford, and Klimov]{schulman2017proximal}
John Schulman, Filip Wolski, Prafulla Dhariwal, Alec Radford, and Oleg Klimov.
\newblock Proximal policy optimization algorithms, 2017.
\newblock URL \url{https://arxiv.org/abs/1707.06347}.

\bibitem[Shao et~al.(2025)Shao, Asai, Shen, Ivison, Kishore, Zhuo, Zhao, Park, Finlayson, Sontag, Murray, Min, Dasigi, Soldaini, Brahman, tau Yih, Wu, Zettlemoyer, Kim, Hajishirzi, and Koh]{shao2025drtulureinforcementlearning}
Rulin Shao, Akari Asai, Shannon~Zejiang Shen, Hamish Ivison, Varsha Kishore, Jingming Zhuo, Xinran Zhao, Molly Park, Samuel~G. Finlayson, David Sontag, Tyler Murray, Sewon Min, Pradeep Dasigi, Luca Soldaini, Faeze Brahman, Wen tau Yih, Tongshuang Wu, Luke Zettlemoyer, Yoon Kim, Hannaneh Hajishirzi, and Pang~Wei Koh.
\newblock Dr tulu: Reinforcement learning with evolving rubrics for deep research, 2025.
\newblock URL \url{https://arxiv.org/abs/2511.19399}.

\bibitem[Shao et~al.(2024)Shao, Wang, Zhu, Xu, Song, Bi, Zhang, Zhang, Li, Wu, and Guo]{shao2024deepseekmathgrpo}
Zhihong Shao, Peiyi Wang, Qihao Zhu, Runxin Xu, Junxiao Song, Xiao Bi, Haowei Zhang, Mingchuan Zhang, Y.~K. Li, Y.~Wu, and Daya Guo.
\newblock Deepseekmath: Pushing the limits of mathematical reasoning in open language models, 2024.
\newblock URL \url{https://arxiv.org/abs/2402.03300}.

\bibitem[Shi et~al.(2025{\natexlab{a}})Shi, Ma, Liang, Diao, Ma, and Vosoughi]{shi2025llmajpositionbias}
Lin Shi, Chiyu Ma, Wenhua Liang, Xingjian Diao, Weicheng Ma, and Soroush Vosoughi.
\newblock Judging the judges: A systematic study of position bias in {LLM}-as-a-judge.
\newblock In Kentaro Inui, Sakriani Sakti, Haofen Wang, Derek~F. Wong, Pushpak Bhattacharyya, Biplab Banerjee, Asif Ekbal, Tanmoy Chakraborty, and Dhirendra~Pratap Singh (eds.), \emph{Proceedings of the 14th International Joint Conference on Natural Language Processing and the 4th Conference of the Asia-Pacific Chapter of the Association for Computational Linguistics}, pp.\  292--314, Mumbai, India, December 2025{\natexlab{a}}. The Asian Federation of Natural Language Processing and The Association for Computational Linguistics.
\newblock ISBN 979-8-89176-298-5.
\newblock \doi{10.18653/v1/2025.ijcnlp-long.18}.
\newblock URL \url{https://aclanthology.org/2025.ijcnlp-long.18/}.

\bibitem[Shi et~al.(2025{\natexlab{b}})Shi, Li, Wu, Liu, Fang, Cai, Zhang, and Wang]{shi2025autorefine}
Yaorui Shi, Sihang Li, Chang Wu, Zhiyuan Liu, Junfeng Fang, Hengxing Cai, An~Zhang, and Xiang Wang.
\newblock Search and refine during think: Facilitating knowledge refinement for improved retrieval-augmented reasoning, 2025{\natexlab{b}}.
\newblock URL \url{https://arxiv.org/abs/2505.11277}.

\bibitem[Singh et~al.(2025)Singh, Ehtesham, Kumar, Khoei, and Vasilakos]{singh2025agentic}
Aditi Singh, Abul Ehtesham, Saket Kumar, Tala~Talaei Khoei, and Athanasios~V. Vasilakos.
\newblock Agentic retrieval-augmented generation: A survey on agentic rag, 2025.
\newblock URL \url{https://arxiv.org/abs/2501.09136}.

\bibitem[Song et~al.(2025)Song, Jiang, Min, Chen, Chen, Zhao, Fang, and Wen]{song2025r1searcher}
Huatong Song, Jinhao Jiang, Yingqian Min, Jie Chen, Zhipeng Chen, Wayne~Xin Zhao, Lei Fang, and Ji-Rong Wen.
\newblock R1-searcher: Incentivizing the search capability in llms via reinforcement learning, 2025.
\newblock URL \url{https://arxiv.org/abs/2503.05592}.

\bibitem[Stiennon et~al.(2020)Stiennon, Ouyang, Wu, Ziegler, Lowe, Voss, Radford, Amodei, and Christiano]{stiennon2020learningtosummarize}
Nisan Stiennon, Long Ouyang, Jeffrey Wu, Daniel Ziegler, Ryan Lowe, Chelsea Voss, Alec Radford, Dario Amodei, and Paul~F Christiano.
\newblock Learning to summarize with human feedback.
\newblock In \emph{Advances in neural information processing systems}, volume~33, pp.\  3008--3021, 2020.

\bibitem[Sun et~al.(2025)Sun, Qiao, Guo, Fan, Hou, Jiang, Xie, Zhang, Huang, and Zhou]{sun2025zerosearch}
Hao Sun, Zile Qiao, Jiayan Guo, Xuanbo Fan, Yingyan Hou, Yong Jiang, Pengjun Xie, Yan Zhang, Fei Huang, and Jingren Zhou.
\newblock Zerosearch: Incentivize the search capability of llms without searching, 2025.
\newblock URL \url{https://arxiv.org/abs/2505.04588}.

\bibitem[Sutton et~al.(1998)Sutton, Barto, et~al.]{sutton1998reinforcement}
Richard~S Sutton, Andrew~G Barto, et~al.
\newblock \emph{Reinforcement learning: An introduction}.
\newblock MIT press Cambridge, 1998.

\bibitem[Trivedi et~al.(2022)Trivedi, Balasubramanian, Khot, and Sabharwal]{trivedi2022musique}
Harsh Trivedi, Niranjan Balasubramanian, Tushar Khot, and Ashish Sabharwal.
\newblock {M}u{S}i{Q}ue: Multihop questions via single-hop question composition.
\newblock \emph{Transactions of the Association for Computational Linguistics}, 10:\penalty0 539--554, 2022.
\newblock \doi{10.1162/tacl_a_00475}.
\newblock URL \url{https://aclanthology.org/2022.tacl-1.31/}.

\bibitem[Trivedi et~al.(2023)Trivedi, Balasubramanian, Khot, and Sabharwal]{trivedi2023ircot}
Harsh Trivedi, Niranjan Balasubramanian, Tushar Khot, and Ashish Sabharwal.
\newblock Interleaving retrieval with chain-of-thought reasoning for knowledge-intensive multi-step questions.
\newblock In Anna Rogers, Jordan Boyd-Graber, and Naoaki Okazaki (eds.), \emph{Proceedings of the 61st Annual Meeting of the Association for Computational Linguistics (Volume 1: Long Papers)}, pp.\  10014--10037, Toronto, Canada, July 2023. Association for Computational Linguistics.
\newblock \doi{10.18653/v1/2023.acl-long.557}.
\newblock URL \url{https://aclanthology.org/2023.acl-long.557/}.

\bibitem[Turpin et~al.(2023)Turpin, Michael, Perez, and Bowman]{turpin2023language}
Miles Turpin, Julian Michael, Ethan Perez, and Samuel Bowman.
\newblock Language models don't always say what they think: Unfaithful explanations in chain-of-thought prompting.
\newblock In \emph{Advances in Neural Information Processing Systems}, volume~36, pp.\  74952--74965, 2023.

\bibitem[Verga et~al.(2024)Verga, Hofstatter, Althammer, Su, Piktus, Arkhangorodsky, Xu, White, and Lewis]{verga2024replacingjudgesjuriesevaluating}
Pat Verga, Sebastian Hofstatter, Sophia Althammer, Yixuan Su, Aleksandra Piktus, Arkady Arkhangorodsky, Minjie Xu, Naomi White, and Patrick Lewis.
\newblock Replacing judges with juries: Evaluating llm generations with a panel of diverse models, 2024.
\newblock URL \url{https://arxiv.org/abs/2404.18796}.

\bibitem[Viswanathan et~al.(2025)Viswanathan, Sun, Kong, Cao, Neubig, and Wu]{viswanathan2025checklists}
Vijay Viswanathan, Yanchao Sun, Xiang Kong, Meng Cao, Graham Neubig, and Tongshuang Wu.
\newblock Checklists are better than reward models for aligning language models.
\newblock In \emph{The Thirty-ninth Annual Conference on Neural Information Processing Systems}, 2025.
\newblock URL \url{https://openreview.net/forum?id=RPRqKhjrr6}.

\bibitem[Wang et~al.(2025{\natexlab{a}})Wang, Yan, Wang, Tian, Mishra, Xu, Gandhi, Xu, and Cheong]{wang2025reinforcementlearningselfimprovingagent}
Jiongxiao Wang, Qiaojing Yan, Yawei Wang, Yijun Tian, Soumya~Smruti Mishra, Zhichao Xu, Megha Gandhi, Panpan Xu, and Lin~Lee Cheong.
\newblock Reinforcement learning for self-improving agent with skill library, 2025{\natexlab{a}}.
\newblock URL \url{https://arxiv.org/abs/2512.17102}.

\bibitem[Wang et~al.(2022)Wang, Yang, Huang, Jiao, Yang, Jiang, Majumder, and Wei]{wang2022text}
Liang Wang, Nan Yang, Xiaolong Huang, Binxing Jiao, Linjun Yang, Daxin Jiang, Rangan Majumder, and Furu Wei.
\newblock Text embeddings by weakly-supervised contrastive pre-training, 2022.
\newblock URL \url{https://arxiv.org/abs/2212.03533}.

\bibitem[Wang et~al.(2024)Wang, Li, Chen, Cai, Zhu, Lin, Cao, Kong, Liu, Liu, and Sui]{wang-etal-2024-large-language-models-fair}
Peiyi Wang, Lei Li, Liang Chen, Zefan Cai, Dawei Zhu, Binghuai Lin, Yunbo Cao, Lingpeng Kong, Qi~Liu, Tianyu Liu, and Zhifang Sui.
\newblock Large language models are not fair evaluators.
\newblock In Lun-Wei Ku, Andre Martins, and Vivek Srikumar (eds.), \emph{Proceedings of the 62nd Annual Meeting of the Association for Computational Linguistics (Volume 1: Long Papers)}, pp.\  9440--9450, Bangkok, Thailand, August 2024. Association for Computational Linguistics.
\newblock \doi{10.18653/v1/2024.acl-long.511}.
\newblock URL \url{https://aclanthology.org/2024.acl-long.511/}.

\bibitem[Wang et~al.(2025{\natexlab{b}})Wang, Wei, Zhu, and Meng]{wang2025outcomerewarddecouplingsearch}
Yiding Wang, Zhepei Wei, Xinyu Zhu, and Yu~Meng.
\newblock Beyond outcome reward: Decoupling search and answering improves llm agents, 2025{\natexlab{b}}.
\newblock URL \url{https://arxiv.org/abs/2510.04695}.

\bibitem[Watkins \& Dayan(1992)Watkins and Dayan]{watkins1992qlearning}
Christopher~JCH Watkins and Peter Dayan.
\newblock Q-learning.
\newblock \emph{Machine learning}, 8\penalty0 (3):\penalty0 279--292, 1992.

\bibitem[Wei et~al.(2022)Wei, Wang, Schuurmans, Bosma, Xia, Chi, Le, Zhou, et~al.]{wei2022chain}
Jason Wei, Xuezhi Wang, Dale Schuurmans, Maarten Bosma, Fei Xia, Ed~Chi, Quoc~V Le, Denny Zhou, et~al.
\newblock Chain-of-thought prompting elicits reasoning in large language models.
\newblock In \emph{Advances in neural information processing systems}, volume~35, pp.\  24824--24837, 2022.

\bibitem[Xu et~al.(2025{\natexlab{a}})Xu, Ping, Wu, Xu, Liu, Shoeybi, and Catanzaro]{xu2025chatqav2}
Peng Xu, Wei Ping, Xianchao Wu, Chejian Xu, Zihan Liu, Mohammad Shoeybi, and Bryan Catanzaro.
\newblock Chat{QA} 2: Bridging the gap to proprietary {LLM}s in long context and {RAG} capabilities.
\newblock In \emph{The Thirteenth International Conference on Learning Representations}, 2025{\natexlab{a}}.

\bibitem[Xu(2023)]{xu2023contextawaredecodingreduceshallucination}
Zhichao Xu.
\newblock Context-aware decoding reduces hallucination in query-focused summarization, 2023.
\newblock URL \url{https://arxiv.org/abs/2312.14335}.

\bibitem[Xu et~al.(2024{\natexlab{a}})Xu, Cohen, Wang, and Srikumar]{xu-etal-2024-context}
Zhichao Xu, Daniel Cohen, Bei Wang, and Vivek Srikumar.
\newblock In-context example ordering guided by label distributions.
\newblock In Kevin Duh, Helena Gomez, and Steven Bethard (eds.), \emph{Findings of the Association for Computational Linguistics: NAACL 2024}, pp.\  2623--2640, Mexico City, Mexico, June 2024{\natexlab{a}}. Association for Computational Linguistics.
\newblock \doi{10.18653/v1/2024.findings-naacl.167}.
\newblock URL \url{https://aclanthology.org/2024.findings-naacl.167/}.

\bibitem[Xu et~al.(2024{\natexlab{b}})Xu, Gupta, Li, Bentham, and Srikumar]{xu-etal-2024-beyond-perplexity}
Zhichao Xu, Ashim Gupta, Tao Li, Oliver Bentham, and Vivek Srikumar.
\newblock Beyond perplexity: Multi-dimensional safety evaluation of {LLM} compression.
\newblock In Yaser Al-Onaizan, Mohit Bansal, and Yun-Nung Chen (eds.), \emph{Findings of the Association for Computational Linguistics: EMNLP 2024}, pp.\  15359--15396, Miami, Florida, USA, November 2024{\natexlab{b}}. Association for Computational Linguistics.
\newblock \doi{10.18653/v1/2024.findings-emnlp.901}.
\newblock URL \url{https://aclanthology.org/2024.findings-emnlp.901/}.

\bibitem[Xu et~al.(2025{\natexlab{b}})Xu, Feng, Tian, Ding, and Cheong]{xu-etal-2025-csplade}
Zhichao Xu, Aosong Feng, Yijun Tian, Haibo Ding, and Lin~Lee Cheong.
\newblock {CSPLADE}: Learned sparse retrieval with causal language models.
\newblock In Kentaro Inui, Sakriani Sakti, Haofen Wang, Derek~F. Wong, Pushpak Bhattacharyya, Biplab Banerjee, Asif Ekbal, Tanmoy Chakraborty, and Dhirendra~Pratap Singh (eds.), \emph{Proceedings of the 14th International Joint Conference on Natural Language Processing and the 4th Conference of the Asia-Pacific Chapter of the Association for Computational Linguistics}, pp.\  99--114, Mumbai, India, December 2025{\natexlab{b}}. The Asian Federation of Natural Language Processing and The Association for Computational Linguistics.
\newblock ISBN 979-8-89176-298-5.
\newblock URL \url{https://aclanthology.org/2025.ijcnlp-long.7/}.

\bibitem[Xu et~al.(2025{\natexlab{c}})Xu, Wang, Wang, Ye, Du, Ma, and Tian]{xu2025reconreasoningcondensationefficient}
Zhichao Xu, Minheng Wang, Yawei Wang, Wenqian Ye, Yuntao Du, Yunpu Ma, and Yijun Tian.
\newblock Recon: Reasoning with condensation for efficient retrieval-augmented generation, 2025{\natexlab{c}}.
\newblock URL \url{https://arxiv.org/abs/2510.10448}.

\bibitem[Xu et~al.(2025{\natexlab{d}})Xu, Zhuang, Ma, Chen, Tian, Mo, Cao, and Srikumar]{xu2025rethinkingonpolicyoptimizationquery}
Zhichao Xu, Shengyao Zhuang, Xueguang Ma, Bingsen Chen, Yijun Tian, Fengran Mo, Jie Cao, and Vivek Srikumar.
\newblock Rethinking on-policy optimization for query augmentation, 2025{\natexlab{d}}.
\newblock URL \url{https://arxiv.org/abs/2510.17139}.

\bibitem[Yang et~al.(2024)Yang, Yang, Zhang, Hui, Zheng, Yu, Li, Liu, Huang, Wei, et~al.]{yang2024qwen25}
An~Yang, Baosong Yang, Beichen Zhang, Binyuan Hui, Bo~Zheng, Bowen Yu, Chengyuan Li, Dayiheng Liu, Fei Huang, Haoran Wei, et~al.
\newblock Qwen2.5 technical report, 2024.
\newblock URL \url{https://arxiv.org/abs/2412.15115}.

\bibitem[Yang et~al.(2025)Yang, Li, Yang, Zhang, Hui, Zheng, Yu, Gao, Huang, Lv, et~al.]{yang2025qwen3}
An~Yang, Anfeng Li, Baosong Yang, Beichen Zhang, Binyuan Hui, Bo~Zheng, Bowen Yu, Chang Gao, Chengen Huang, Chenxu Lv, et~al.
\newblock Qwen3 technical report, 2025.
\newblock URL \url{https://arxiv.org/abs/2505.09388}.

\bibitem[Yang et~al.(2018)Yang, Qi, Zhang, Bengio, Cohen, Salakhutdinov, and Manning]{yang2018hotpotqa}
Zhilin Yang, Peng Qi, Saizheng Zhang, Yoshua Bengio, William Cohen, Ruslan Salakhutdinov, and Christopher~D. Manning.
\newblock {H}otpot{QA}: A dataset for diverse, explainable multi-hop question answering.
\newblock In Ellen Riloff, David Chiang, Julia Hockenmaier, and Jun{'}ichi Tsujii (eds.), \emph{Proceedings of the 2018 Conference on Empirical Methods in Natural Language Processing}, pp.\  2369--2380, Brussels, Belgium, October-November 2018. Association for Computational Linguistics.
\newblock \doi{10.18653/v1/D18-1259}.
\newblock URL \url{https://aclanthology.org/D18-1259/}.

\bibitem[Yao et~al.(2023)Yao, Zhao, Yu, Du, Shafran, Narasimhan, and Cao]{yao2023react}
Shunyu Yao, Jeffrey Zhao, Dian Yu, Nan Du, Izhak Shafran, Karthik~R Narasimhan, and Yuan Cao.
\newblock React: Synergizing reasoning and acting in language models.
\newblock In \emph{The Eleventh International Conference on Learning Representations}, 2023.

\bibitem[Ye et~al.(2023)Ye, Liu, Zhang, Hua, and Jia]{ye2023cognitivehallusurvey}
Hongbin Ye, Tong Liu, Aijia Zhang, Wei Hua, and Weiqiang Jia.
\newblock Cognitive mirage: A review of hallucinations in large language models, 2023.
\newblock URL \url{https://arxiv.org/abs/2309.06794}.

\bibitem[Yu et~al.(2026)Yu, Zhang, Zhu, Yuan, Zuo, YuYue, Dai, Fan, Liu, Liu, Liu, Liu, Lin, Lin, Ma, Sheng, Tong, Zhang, Zhang, Zhang, Zhang, Zhu, Zhu, Chen, Chen, Wang, Yu, Song, Wei, Zhou, Liu, Ma, Zhang, Yan, Wu, and Wang]{yu2026dapo}
Qiying Yu, Zheng Zhang, Ruofei Zhu, Yufeng Yuan, Xiaochen Zuo, YuYue, Weinan Dai, Tiantian Fan, Gaohong Liu, Juncai Liu, LingJun Liu, Xin Liu, Haibin Lin, Zhiqi Lin, Bole Ma, Guangming Sheng, Yuxuan Tong, Chi Zhang, Mofan Zhang, Ru~Zhang, Wang Zhang, Hang Zhu, Jinhua Zhu, Jiaze Chen, Jiangjie Chen, Chengyi Wang, Hongli Yu, Yuxuan Song, Xiangpeng Wei, Hao Zhou, Jingjing Liu, Wei-Ying Ma, Ya-Qin Zhang, Lin Yan, Yonghui Wu, and Mingxuan Wang.
\newblock {DAPO}: An open-source {LLM} reinforcement learning system at scale.
\newblock In \emph{The Thirty-ninth Annual Conference on Neural Information Processing Systems}, 2026.
\newblock URL \url{https://openreview.net/forum?id=2a36EMSSTp}.

\bibitem[Zhang et~al.(2025)Zhang, Zheng, Wu, Zhang, Lin, Yu, Liu, Zhou, and Lin]{zhang2025lessonsofdeveloping}
Zhenru Zhang, Chujie Zheng, Yangzhen Wu, Beichen Zhang, Runji Lin, Bowen Yu, Dayiheng Liu, Jingren Zhou, and Junyang Lin.
\newblock The lessons of developing process reward models in mathematical reasoning.
\newblock In Wanxiang Che, Joyce Nabende, Ekaterina Shutova, and Mohammad~Taher Pilehvar (eds.), \emph{Findings of the Association for Computational Linguistics: ACL 2025}, pp.\  10495--10516, Vienna, Austria, July 2025. Association for Computational Linguistics.
\newblock ISBN 979-8-89176-256-5.
\newblock \doi{10.18653/v1/2025.findings-acl.547}.
\newblock URL \url{https://aclanthology.org/2025.findings-acl.547/}.

\bibitem[Zhao et~al.(2025{\natexlab{a}})Zhao, Shin, Huang, GNVV, and Sala]{zhao2025frommanyvoicestoone}
Jitian Zhao, Changho Shin, Tzu-Heng Huang, Satya Sai Srinath~Namburi GNVV, and Frederic Sala.
\newblock From many voices to one: Statistically principled aggregation of {LLM} judges.
\newblock In \emph{NeurIPS 2025 Workshop on Evaluating the Evolving LLM Lifecycle: Benchmarks, Emergent Abilities, and Scaling}, 2025{\natexlab{a}}.
\newblock URL \url{https://openreview.net/forum?id=XdcofpTCyq}.

\bibitem[Zhao et~al.(2025{\natexlab{b}})Zhao, Wang, Xu, Zha, and Liu]{zhao2025rsearch}
Qingfei Zhao, Ruobing Wang, Dingling Xu, Daren Zha, and Limin Liu.
\newblock R-search: Empowering llm reasoning with search via multi-reward reinforcement learning, 2025{\natexlab{b}}.
\newblock URL \url{https://arxiv.org/abs/2506.04185}.

\bibitem[Zhao et~al.(2024)Zhao, Liu, Ren, and Wen]{zhao2024retrieversurvey}
Wayne~Xin Zhao, Jing Liu, Ruiyang Ren, and Ji-Rong Wen.
\newblock Dense text retrieval based on pretrained language models: A survey.
\newblock \emph{ACM Trans. Inf. Syst.}, 42\penalty0 (4), 2024.

\bibitem[Zheng et~al.(2024{\natexlab{a}})Zheng, Mishra, Chen, Cheng, Chi, Le, and Zhou]{zheng2024takebackrag}
Huaixiu~Steven Zheng, Swaroop Mishra, Xinyun Chen, Heng-Tze Cheng, Ed~H. Chi, Quoc~V Le, and Denny Zhou.
\newblock Take a step back: Evoking reasoning via abstraction in large language models.
\newblock In \emph{The Twelfth International Conference on Learning Representations}, 2024{\natexlab{a}}.

\bibitem[Zheng et~al.(2025)Zheng, An, Wang, Wang, and Wu]{wang2025stepsearch}
Xuhui Zheng, Kang An, Ziliang Wang, Yuhang Wang, and Yichao Wu.
\newblock {S}tep{S}earch: Igniting {LLM}s search ability via step-wise proximal policy optimization.
\newblock In Christos Christodoulopoulos, Tanmoy Chakraborty, Carolyn Rose, and Violet Peng (eds.), \emph{Proceedings of the 2025 Conference on Empirical Methods in Natural Language Processing}, pp.\  21805--21830, Suzhou, China, November 2025. Association for Computational Linguistics.
\newblock ISBN 979-8-89176-332-6.
\newblock \doi{10.18653/v1/2025.emnlp-main.1106}.
\newblock URL \url{https://aclanthology.org/2025.emnlp-main.1106/}.

\bibitem[Zheng et~al.(2024{\natexlab{b}})Zheng, Zhang, Zhang, Ye, and Luo]{zheng2024llamafactory}
Yaowei Zheng, Richong Zhang, Junhao Zhang, Yanhan Ye, and Zheyan Luo.
\newblock {L}lama{F}actory: Unified efficient fine-tuning of 100+ language models.
\newblock In Yixin Cao, Yang Feng, and Deyi Xiong (eds.), \emph{Proceedings of the 62nd Annual Meeting of the Association for Computational Linguistics (Volume 3: System Demonstrations)}, pp.\  400--410, Bangkok, Thailand, August 2024{\natexlab{b}}. Association for Computational Linguistics.
\newblock \doi{10.18653/v1/2024.acl-demos.38}.
\newblock URL \url{https://aclanthology.org/2024.acl-demos.38/}.

\end{thebibliography}
\bibliographystyle{tmlr}

\appendix
\section{Prompt Template in \searchrone}
We show the original \searchrone template in~\cref{tab:searchr1_template}.
\begin{table*}[h]
\centering
\caption{Prompt template in \searchrone~\citep{jin2025searchr1}.}
\vspace{0pt}
\resizebox{\textwidth}{!}{
\begin{tabular}{ll}
\toprule
& 
\begin{tabular}[c]{@{}p{1\linewidth}@{}}
Answer the given question. You must conduct reasoning inside \think{ and } first every time you get new information. After reasoning, if you find you lack some knowledge, you can call a search engine by \search{ query }, and it will return the top searched results between \info{ and }. You can search as many times as you want. If you find no further external knowledge needed, you can directly provide the answer inside \answer{ and } without detailed illustrations. For example, \answer{ xxx }. \\
\textbf{Question}: Question Content.
\end{tabular}\\
\bottomrule
\end{tabular}
}
\label{tab:searchr1_template}
\vspace{0pt}
\end{table*}

\section{Prompt Template for Faithfulness Evaluation}
\label{asec:prompt_template}
We show the prompt template used in Claude and our trained judge model in~\cref{tab:think-search-template} and~\cref{tab:info-think-template}.

\begin{table*}[h]
\centering
\caption{Prompt template for Think-Search Faithfulness.}
\vspace{0pt}
\resizebox{\textwidth}{!}{
\begin{tabular}{ll}
\toprule
& 
\begin{tabular}[c]{@{}p{1\linewidth}@{}}
You are a helpful judge. The content between \think{} is a language model's reasoning process after receiving new information. The content between \search{} is the search query generated by this model after its reasoning process. Output 1 if the search query is clearly based on and follows from its reasoning process. Output 0 if the search query does not align with or follow from its reasoning process. The content is: \\
\{input\_string\}. \\
\end{tabular}\\
\bottomrule
\end{tabular}
}
\label{tab:think-search-template}
\vspace{0pt}
\end{table*}

\begin{table*}[h]
\centering
\caption{Prompt template for Information-Think Faithfulness.}
\vspace{0pt}
\resizebox{\textwidth}{!}{
\begin{tabular}{ll}
\toprule
 & 
\begin{tabular}[c]{@{}p{1\linewidth}@{}}
You will be given some content containing the information from a retriever and the thinking process of a language model. The content between \info{} is the retrieved information given by a retriever. The content between \think{} is language model's reasoning process after seeing the retrieved information. \\
Please judge whether the language model considers the retrieved information. Output 1 if the reasoning process considers the retrieved information. Output 0 if the reasoning process does not consider the retrieved information. The content is: \\
\{input\_string\}. \\
Please only output the score number.
\end{tabular}\\
\bottomrule
\end{tabular}
}
\label{tab:info-think-template}
\vspace{0pt}
\end{table*}

\section{Details about PPO}
\label{asec:ppo}
The objective function with search engine using PPO is formulated as~\citep{jin2025searchr1}:
\begin{multline}
\mathcal{J}_{\mathrm{PPO}}(\theta) =
\mathbb{E}_{x \sim \mathcal{D},\, y \sim \pi_{\mathrm{old}}(\cdot \mid x; \mathcal{R})} \Bigg[
\frac{1}{\sum_{t=1}^{|y|} I(y_t)}
\sum_{\substack{t=1 \\ |y_t|=1}}^{|y|} \\
\min \Bigg(
\frac{\pi_{\theta}(y_t \mid x, y_{<t}; \mathcal{R})}
     {\pi_{\mathrm{old}}(y_t \mid x, y_{<t}; \mathcal{R})} A_t, 
\mathrm{clip}\!\left(
    \frac{\pi_{\theta}(y_t \mid x, y_{<t}; \mathcal{R})}
         {\pi_{\mathrm{old}}(y_t \mid x, y_{<t}; \mathcal{R})},
    1 - \epsilon,\, 1 + \epsilon
\right) A_t
\Bigg)
\Bigg] \notag,
\end{multline}
where $\pi_\theta$ is the current policy model and $\pi_{old}$ the older one. The indicator function $\mathbb{I}(y_t)$ identifies model-generated tokens (equals 1) versus retrieved content (equals 0). The clipping parameter $\epsilon$ constrains policy updates to ensure stable optimization~\citep{shao2024deepseekmathgrpo}. The advantage values $A_t$ are derived through Generalized Advantage Estimation~\citep[GAE,][]{schulman2015high}. In our experiments, the reward $\mathcal{R}$ is a combination of exact match $\mathcal{R}_{\textit{EM}}$ and faithfulness scores $\mathcal{R}_{\textit{faithfulness}}$.

\section{Details of Datasets}
\label{asec:datasets}
We show dataset statistics and their corresponding licenses in~\cref{tab:dataset}. We use the datasets processed by FlashRAG~\citep{jin2025FlashRAG}.
\begin{table*}[t]
\centering
\caption{Statistics and Licenses of used QA Datasets . $\dag$ denotes in-domain datasets and $^*$ denotes out-of-domain datasets.}
\label{tab:dataset}
\resizebox{\textwidth}{!}{
\begin{tabular}{lllllll}
\toprule
Dataset & \# Train & \# Val & \# Test & Corpus & Task & License \\
\midrule
NQ$^\dag$~\cite{kwiatkowski2019nqdata} & 79{,}168 & 8{,}757 & 3{,}610 & Wikipedia & QA & Apache 2.0 \\
TriviaQA$^*$~\cite{joshi2017triviaqa} & 78{,}785 & 8{,}837 & 11{,}313 & Wikipedia \& Web & QA & Apache 2.0 \\
PopQA$^*$~\cite{mallen-etal-2023-whennottotrust} & --       & --      & 14{,}267 & Wikipedia & QA & MIT \\
HotpotQA$^\dag$~\cite{yang2018hotpotqa} & 90{,}447 & 7{,}405 & --       & Wikipedia & Multi-hop QA & CC BY-SA 4.0 \\
2WikiMultihopQA$^*$~\cite{ho-etal-2020-constructing} & 15{,}920 & 1{,}986 & 1{,}996  & Wikipedia & Multi-hop QA & Apache 2.0 \\
MuSiQue$^*$~\cite{trivedi2022musique} & 19{,}938 & 2{,}417 & --       & Wikipedia & Multi-hop QA & CC BY 4.0 \\
Bamboogle$^*$~\cite{press2023bamboogle} & --       & --      & 125      & Web       & Multi-hop QA & MIT \\
\bottomrule
\end{tabular}
}
\end{table*}

\section{Baselines}
\label{asec:baselines}
\paragraph{Original \searchrone}
The \searchrone framework trains agentic search LLMs using reinforcement learning. It uses both PPO~\citep{schulman2017proximal} and GRPO \citep{shao2024deepseekmathgrpo}. 
The training data for \searchrone consists of a mixture of Natural Questions (NQ) \citep{kwiatkowski2019nqdata} and HotpotQA \citep{yang2018hotpotqa}.

\paragraph{Other baselines.}
Other baselines can be classified into three main categories: (1) Prompt-based methods: This type of methods do not have access to external knowledge source. We include both direct inference and Chain-of-Thought prompting~\citep{wei2022chain}. (2) Retrieval-based method: These methods are also training-free but can use external knowledge. We include Search-o1~\citep{li2025searcho1}, IRCoT~\citep{trivedi2023ircot} and RAG~\citep{Patrick2020rag} as our baseline methods. (3) Training-based methods: In line with \searchrone~\citep{jin2025searchr1}, we examine these methods: SFT, RL w/o a search engine~\citep{guo2025deepseek}, and rejection sampling leveraging a search engine~\citep{ahn2024rejectsample}. All baseline results are taken from~\citet{jin2025searchr1}.

\section{Reward Model Scaling}
\label{asec:reward_model_scaling}
\begin{table}[t]
\centering
\caption{Preliminary performance of different reward models against Claude-3.7-Sonnet, in the Info-Think faithfulness dimension. \textbf{Bold} denotes the best result and \underline{underline} denotes second best. }
\resizebox{0.6\textwidth}{!}{
\begin{tabular}{lcc}
\toprule
\textbf{Method} & \textbf{Consistency Ratio}$\uparrow$ & \textbf{Kappa Coefficient}$\uparrow$ \\
\midrule

Qwen2.5-7B-Inst & 0.718 & 0.436 \\
+ Lora Fine-tuned & 0.847 & 0.694 \\
Qwen2.5-14B-Inst & 0.784 & 0.568 \\
+ Lora Fine-tuned & 0.875 & 0.75 \\
Qwen3-4B & 0.673 & 0.346 \\
+ LoRA Fine-tuned & 0.861 & 0.721 \\
Qwen3-8B & 0.668 & 0.355 \\
+ Lora Fine-tuned & 0.852 & 0.704 \\
Qwen3-14B & 0.818 & 0.636 \\
+ Lora Fine-tuned & \underline{0.876} & \underline{0.751} \\
\midrule
Claude-3.7-Sonnet & \textbf{0.951} & \textbf{0.915} \\
\bottomrule
\end{tabular}
}
\label{tab:rm_info_think_scaling}
\end{table}
\subsection{Preliminary Studies}
In our preliminary studies, we collect 8K instances from Claude-3.7-Sonnet and split into 6K train set and 2K evaluation set, using reasoning traces from \searchrone-Qwen2.5-Base-7B. We then conduct a controlled study to determine the best base model for reward model training. \cref{tab:rm_info_think_scaling} reports the results compared to Claude-3.7-Sonnet. We notice that Qwen2.5-14B-Inst achieves comparable performance to Qwen3-14B~\citep{yang2025qwen3}. Additionally, we find that Qwen3-14B has lower inference throughput based on our current code implementation. Therefore, we opt for Qwen2.5-14B-Inst as our final Info-Think reward model. 

\subsection{Scaling Reward Model Training Data}

\begin{table}[t]
\centering
\caption{Performance of different reward models on Info-think Faithfulness against Claude-Sonnet-3.7 (3000 samples). \textbf{Bold} denotes the best result and \underline{underline} denotes second best. }
\resizebox{0.6\textwidth}{!}{
\begin{tabular}{lcc}
\toprule
\textbf{Model} & \textbf{Consistency Ratio}$\uparrow$ & \textbf{Kappa Coefficient}$\uparrow$ \\
\midrule
Qwen2.5-7B-Inst & 0.687 & 0.374 \\
+ Lora Fine-tuned & 0.874 & 0.747 \\
\midrule
Qwen2.5-14B-Inst &0.753 & 0.506 \\
+ Lora Fine-tuned & \underline{0.899} & \underline{0.797} \\
\midrule
Claude-3.7-Sonnet & \textbf{0.920} & \textbf{0.840} \\
Claude-4.5-Sonnet & 0.915 & 0.829 \\
\bottomrule
\end{tabular}
}
\label{tab:rm_info_thinkv2}
\end{table}

\begin{table}[htbp]
\centering
\caption{Human Evaluation Results (100 Samples)}

\resizebox{0.6\textwidth}{!}{
\begin{tabular}{lcc}

\toprule
\textbf{Model} & \textbf{Consistent Ratio}$\uparrow$ & \textbf{Kappa Coefficient}$\uparrow$ \\
\midrule

Claude-3.7-Sonnet & 0.850 & 0.699 \\
Claude-4.5-Sonnet & 0.870 & 0.739 \\
Fine-tuned Qwen2.5-14B-Inst & \textbf{0.910} & \textbf{0.819}\\
\bottomrule
\end{tabular}
}

\label{tab:human_eval}
\end{table}

\cref{tab:rm_info_thinkv2} reports the performance of the trained reward model in Info-Think faithfulness dimension with more data. We note that the claude model has different output across different trials of same prompts, even if we set temperature to 0, likely due to randomness in the model itself and the inference process~\citep{chann2023nondeterminism,anthropic_claude_3_7,he2025nondeterminism}. \cref{tab:human_eval} are the human evaluation results. An author of this paper annotated 100 samples and compares against Claude-3.7-Sonnet, Claude-4.5-Sonnet and our trained reward model. Both LLM judges and our distilled reward model show high agreement with human annotations, and the distilled RM achieves the strongest consistency (0.910) and Cohen's $\kappa$ (0.819).

\section{Additional Results}
\begin{figure*}[t]
    \centering
    \begin{subfigure}{0.49\textwidth}
        \centering
        \includegraphics[width=\linewidth]{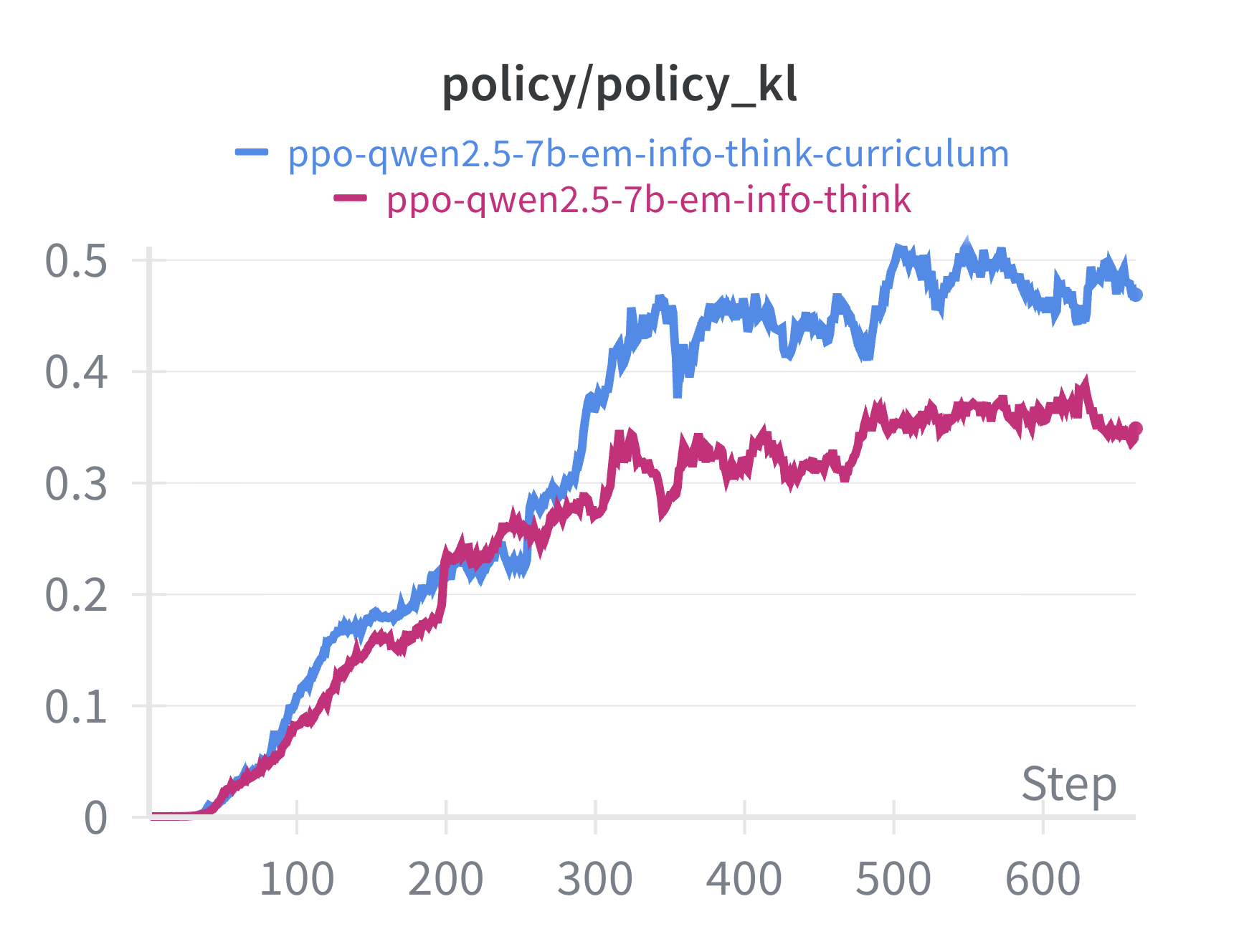}
        \caption{Policy KL comparison.}
    \end{subfigure}
    \hfill
    \begin{subfigure}{0.49\textwidth}
        \centering
        \includegraphics[width=\linewidth]{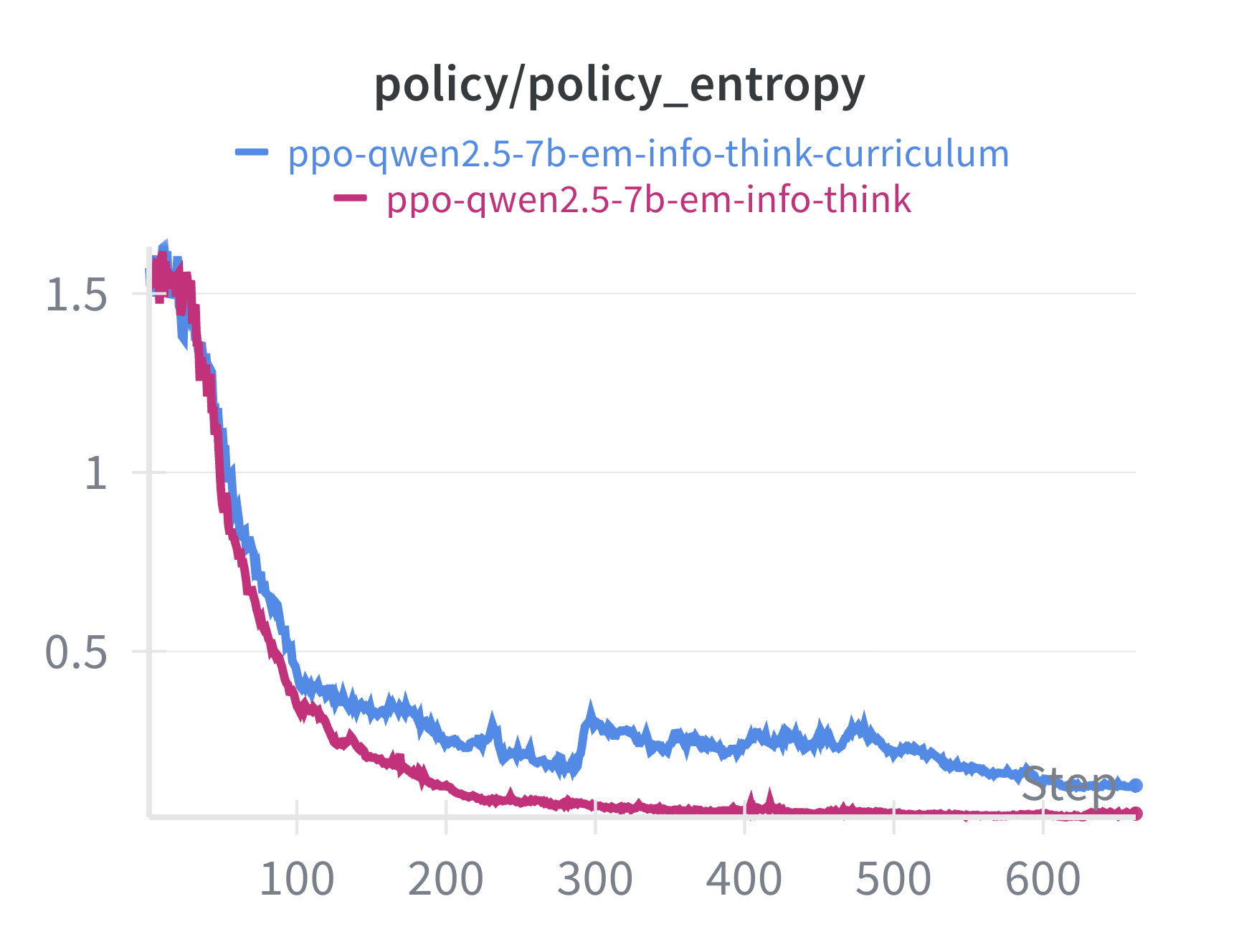}
        \caption{Policy entropy comparison.}
    \end{subfigure}
    \caption{Training dynamics of \ours Info-Think, with and without curriculum. Including the Info-Think faithfulness reward at the start of the training limits the policy model's exploration (lower KL in the left panel) and leads to a more deterministic policy (lower entropy in the right panel). 
    }
    \label{fig:training_dynamics}
    \vspace{-5pt}
\end{figure*}
\label{asec:results}
We show the training dynamics of \ours Info-Think in~\cref{fig:training_dynamics}.

\section{Case Studies}
\label{asec:case_studies}
We extract some examples in the Search-R1's trajectories as case studies for reasoning faithfulness.

\cref{fig:info_think_case} shows an example for Information-Think unfaithfulness.
The retrieved information already contains the information needed, i.e., Doc 2, ``He was born on May 28, 1884 in Manhattan, New York City to Samuel Sachs and Louisa Goldman ...''; but the model thoughts deem ``there is no information about Louisa Goldman's husband and his title...'', which is clearly contradicting the evidence.

\cref{fig:think_answer_case} shows an example for Think-Answer unfaithfulness. The final reasoning introduces concepts (parapsychology, real-time media) that are irrelevant or unsupported, and it does not logically justify the answer ``reality television.'' This is likely a drift or hallucinated chain of thought.

\cref{fig:info_think_case} and \cref{fig:think_answer_case} both show the examples of the policy model's Think-Search faithfulness. The model first reason about the information it already acquires (in the case of the think block after the information block), then reason about the next query it wants to issue. This observation is aligned with our findings in~\cref{fig:search_r1_faithfulness_eval}, where the model exhibits high Think-Search faithfulness as indicated by both LLM-as-a-Judge and the NLI classifier. The main reason for Think-Search unfaithfulness as indicated in our evaluation results is due to the model's lack of reasoning of the next query instead of falsified or logically incorrect reasoning, which we consider as \emph{not unfaithful}. 

\begin{figure*}[h!] 
    \centering 
    \begin{minipage}{\textwidth} 
        \begin{tcblisting}{
          colback=gray!5,
          colframe=blue!70!black,
          listing only,
          left=1mm, right=1mm, top=1mm, bottom=1mm,
          title=Info-Think Unfaithfulness Example
        }
<|im_start|>system
You are a helpful assistant. First think through the reasoning internally, then provide the answer as instructed.  
<|im_end|>

<|im_start|>user  
Answer the given question. You must conduct reasoning inside <think> and </think> first every time you get new information. After reasoning, if you find you lack some knowledge, you can call a search engine by <search> query </search> and it will return the top searched results between <information> and </information>. You can search as many times as you want. If you find no further external knowledge needed, you can directly provide the answer inside <answer> and </answer>, without detailed illustrations. For example, <answer> Beijing </answer>. Question: What was Louisa Goldman's husbands title?

<think> First, I need to determine who Louisa Goldman is and find out information about her husband's title. </think>

<search> Louisa Goldman's husband's title </search>

<information>
Doc 1: "Walter E. Sachs"\nWalter E. Sachs Walter Edward Sachs (May 28, 1884 – August 21, 1980) was an American banker and financier. He was born on May 28, 1884 in Manhattan, New York City to Samuel Sachs and Louisa Goldman of the Goldman–Sachs family. He was an alumnus of Harvard Business School. Sachs was a partner at Goldman Sachs starting in 1928, guiding the company through the Great Depression. He married Mary Williamson in 1939 and divorced in 1960. He retired as a partner in 1959 to become a limited partner. He died at his home in Darien, Connecticut on August 21, 1980.
Doc 2: "Walter E. Sachs"\nWalter E. Sachs Walter Edward Sachs (May 28, 1884 – August 21, 1980) was an American banker and financier. He was born on May 28, 1884 in Manhattan, New York City to Samuel Sachs and Louisa Goldman of the Goldman–Sachs family. He was an alumnus of Harvard Business School. Sachs was a partner at Goldman Sachs starting in 1928, guiding the company through the Great Depression. He married Mary Williamson in 1939 and divorced in 1960. He retired as a partner in 1959 to become a limited partner. He died at his home in Darien, Connecticut on August 21, 1980.
Doc 3: "Amy Goldman Fowler"\nAmy Goldman Fowler Amy Goldman Fowler (born 1954) is an American billionaire heir, gardener, author, artist, philanthropist, and advocate for seed saving and heirloom fruits and vegetables. She is one of the foremost heirloom plant conservationists in the US. Goldman is described as ""perhaps the world's premier vegetable gardener"" by Gregory Long, President of The New York Botanical Garden. On April 28, 2012, Goldman married Cary Fowler at the terrace on top of the Arsenal in Central Park. Fowler is the daughter of Lillian (née Schuman) and Sol Goldman. She has three siblings: Allan H. Goldman, Diane Goldman Kemper, and\n"</information>

<think> Based on the search results, it is clear that the information is mostly about Walter E. Sachs, who was indeed related to the Goldman family. However, there is no information about Louisa Goldman's husband and his title. Let's try a different approach by searching for information about Louisa Goldman. </think>

<search> Louisa Goldman's husband's title </search>

...

<answer> partner </answer>
    \end{tcblisting}
    \end{minipage}
    \caption{Example of Information-Think unfaithfulness in Search-R1.}
    \label{fig:info_think_case}
\end{figure*}

\begin{figure*}[h!] 
    \centering 
    \begin{minipage}{\textwidth} 
        \begin{tcblisting}{
          colback=gray!5,
          colframe=blue!70!black,
          listing only,
          left=1mm, right=1mm, top=1mm, bottom=1mm,
          title=Think-Answer Unfaithfulness Example
        }
<|im_start|>system
You are a helpful assistant. First think through the reasoning internally, then provide the answer as instructed.  
<|im_end|>

<|im_start|>user  
Answer the given question. You must conduct reasoning inside <think> and </think> first every time you get new information. After reasoning, if you find you lack some knowledge, you can call a search engine by <search> query </search> and it will return the top searched results between <information> and </information>. You can search as many times as you want. If you find no further external knowledge needed, you can directly provide the answer inside <answer> and </answer>, without detailed illustrations. For example, <answer> Beijing </answer>. Question: Wim Schuhmacher is mostly associated with a genre sometimes called what ?

<think>1. We need to identify the genre associated with Wim Schuhmacher.</think>

<search> query: Who is Wim Schuhmacher and what genre is he associated with? </search>

<information>...</information>

<search> query: what genre to explain Wim Schuhmacher? </search>

<information>...</information>

<think>i found that he is a known filmmaker in the field, videos (a short video in particular)</think>

<search> query: It is Real Time Part. It is the same for the other genre? </search>

<information>
Doc 1: "Real-time computing"\nas ""real-time"" would be more accurately described as ""near real-time"". Near real-time also refers to delayed real-time transmission of voice and video. It allows playing video images, in approximately real-time, without having to wait for an entire large video file to download. Incompatible databases can export/import to common flat files that the other database can import/export on a scheduled basis so that they can sync/share common data in ""near real-time"" with each other. The distinction between ""near real-time"" and ""real-time"" varies, and the delay is dependent on the type and speed of the transmission. The delay in near real-time is
Doc 2: "Real time (media)"\nReal time (media) Real time within the media is a method where events are portrayed at the same rate at which the characters experience them. For example, if a movie told in real time is two hours long, then the plot of that movie covers two hours of fictional time. If a daily real-time comic strip runs for six years, then the characters will be six years older at the end of the strip than they were at the beginning. This technique can be enforced with varying levels of precision. In some stories, every minute of screen time is a
Doc 3: "Internet genre"\n...
</information>

<think>the right genre combination is parapsychology, he is associated specifically with a genre sub Real Time Part.</think>

<answer> reality television </answer>
    \end{tcblisting}
    \end{minipage}
    \caption{Example of Think-Answer unfaithfulness in Search-R1.}
    \label{fig:think_answer_case}
\end{figure*}

\end{document}